\documentclass{article}

\PassOptionsToPackage{numbers, compress}{natbib}

\usepackage[preprint]{neurips_2026}
\usepackage{graphicx}

\usepackage[utf8]{inputenc} 
\usepackage[T1]{fontenc}    
\usepackage{url}            
\usepackage{booktabs}       
\usepackage{amsfonts}       
\usepackage{nicefrac}       
\usepackage{microtype}      
\usepackage{xcolor}         
\usepackage{amsmath}
\usepackage{multirow}
\usepackage{wrapfig}
\usepackage{float}
\definecolor{lightblue}{HTML}{0071bc}
\usepackage{etoolbox}
\usepackage[colorlinks, citecolor=lightblue, linkcolor=lightblue]{hyperref} 

\usepackage{caption}
\usepackage{enumitem}
\usepackage{subcaption}
\usepackage[utf8]{inputenc}
\usepackage[T1]{fontenc}
\usepackage{booktabs}
\usepackage{longtable}
\usepackage{array}
\usepackage{wrapfig}

\newcommand{\stdv}[1]{\,{\scriptsize\textpm\,#1}}

\title{CLEF: EEG Foundation Model \\ for Learning Clinical Semantics}

\author{%
  Peng Cao\textsuperscript{*}$^1$
  \quad Ali Mirzazadeh\textsuperscript{*}$^1$ \\
  \quad \textbf{Jong Woo Lee}$^2$
  \quad \textbf{Aleksandar Videnovic}$^3$
  \quad \textbf{Dina Katabi} $^1$ \\
  \\
  $^1$ MIT CSAIL \quad $^2$ Brigham and Women's Hospital, Harvard Medical School \\ $^3$ Massachusetts General Hospital, Harvard Medical School
}

\begin{document}
\maketitle

\renewcommand{\thefootnote}{\fnsymbol{footnote}}
\footnotetext[1]{Equal contribution, ordered by coin flip. Correspondence to \texttt{\{alimirz,pengcao\}@mit.edu}.}
\renewcommand{\thefootnote}{\arabic{footnote}}
\setcounter{footnote}{0}

\begin{abstract}
Clinical EEG interpretation requires reasoning over full EEG sessions and integrating signal patterns with clinical context. Existing EEG foundation models are largely designed for short-window decoding and do not incorporate clinical context. We introduce CLEF, a clinically grounded long-context EEG foundation model. CLEF represents EEG sessions as 3D multitaper spectrogram tokens, enabling tractable Transformer modeling at session scale, and aligns embeddings with neurologist reports and structured EHR data through contrastive objectives. We evaluate CLEF on a new 234-task benchmark spanning disease phenotypes, medication exposures, and EEG findings, with more than 260k EEG sessions from over 108k patients. CLEF outperforms prior EEG foundation models on 229 of 234 tasks, improving mean AUROC from 0.65 to 0.74. Reconstruction-only pretraining surpasses prior EEG foundation models, while report and EHR alignment yields further gains. Held-out concept and external-cohort experiments suggest that these representations transfer beyond observed alignment targets. These results support session-scale, clinically grounded representation learning as a promising foundation-model paradigm for clinical EEG.
\end{abstract}
\section{Introduction}
\vspace{-2mm}
\label{sec:intro}
Electroencephalography (EEG) is a cornerstone of clinical neuroscience. By recording electrical activity from the scalp, EEG provides a non-invasive window into brain function and is routinely used to diagnose epilepsy, characterize encephalopathy, monitor neurological deterioration, assess medication effects, and develop biomarkers for neurological and psychiatric diseases. Over one million clinical EEGs are performed annually in the US, yet up to 75\% are interpreted by clinicians without neurophysiology training~\citep{sun2025harvard}. This gap motivates automated systems that extract clinically meaningful information from EEG and democratize expert-level neurophysiological assessment.

Though recent years have seen rapid progress in EEG foundation models~\citep{kostas2021bendr,chien2022maeeg,cui2024neuro,jiang2024large,chen2024eegformer,wang2025cbramod,wang2024eegpt,zhou2025csbrain,yang2025thdbar,ouahidi2025reve}, these models are motivated by brain--computer interface (BCI) research. Their goal is to decode an instantaneous brain state (e.g., a motor intention, an emotional response, a perceived stimulus) from a brief window of neural activity. Accordingly, they typically operate on short EEG segments, often 5--30 seconds long, and learn representations from the EEG signal itself through masked reconstruction, contrastive learning, autoregressive prediction, or related self-supervised objectives. This paradigm has been highly effective for short-window decoding. However, it is not the paradigm used in clinical EEG.

Clinical EEG interpretation is fundamentally different. Neurologists do not interpret an EEG from a 10-second segment; they examine an entire recording session, typically 20 minutes or longer, and integrate patterns that recur, persist, evolve, or fluctuate over time~\citep{sinha2016american}. Epileptiform abnormalities are assessed not only by the morphology of individual discharges, but also by their recurrence and evolution. Encephalopathy, medication effects, and systemic illness often appear as broader changes in background organization, slowing, periodicity, or sleep-wake structure rather than as isolated local waveforms. Clinical EEG is also interpreted in context, incorporating the indication for the study, medications, diagnoses, and broader medical history~\citep{tatum2016american}. Accordingly, a foundation model for clinical EEG should represent an entire session in its clinical context, rather than only short segments.

We introduce \textbf{CLEF}, a \textbf{CL}inical \textbf{E}EG \textbf{F}oundation model built on two principles: \emph{long-context modeling} at the scale of a full EEG session, and \emph{grounding in clinical semantics}. Extending EEG foundation models from seconds to full sessions is not merely increasing context length. A standard 19-channel, 200 Hz, 1280-second recording contains nearly five million raw samples, making direct long-context waveform modeling computationally expensive. Furthermore, modeling EEG waveforms directly over long windows is statistically ill-posed since ongoing EEG is well-described as a non-stationary stochastic process whose instantaneous phase carries little reproducible information across repeated recordings, while its second-order spectral structure is comparatively stable and is what neurologists read off the trace~\cite{cohen2014analyzing, buzsaki2006rhythms, mitra2007observed}. Two recordings from the same patient under identical conditions share spectral content but differ entirely in instantaneous waveform shape. CLEF resolves these issues by operating on multitaper spectrograms~\citep{thomson1982spectrum}, which discard phase and encode the spectral features. To manage the resulting data volume CLEF tokenizes the multi-channel spectrogram via a VQGAN~\citep{esser2021taming} that encodes all channels jointly, compressing a full session into 2{,}048 tokens, and making session-scale Transformer modeling tractable.

Long-context modeling alone cannot recover important clinical semantics. Whether a patient carries a diagnosis of Alzheimer's disease or is being treated with morphine produces signatures in the EEG, but recovering them requires grounding in the clinical modalities that encode this information. CLEF aligns the recording-level embedding with two such modalities through symmetric contrastive objectives~\citep{radford2021learning}: free-text neurologist reports, preprocessed by an LLM summarizer to preserve electrographic content, and structured electronic health record (EHR) data,  which summarizes the patient’s broader clinical profile through demographics, active medications, and diagnoses encoded as a variable-length set of learned code embeddings. Together, reports and EHRs ensure EEG representations are grounded in the clinical semantics used by physicians.

\begin{figure}[t]
    \centering
    \includegraphics[width=0.95\linewidth]{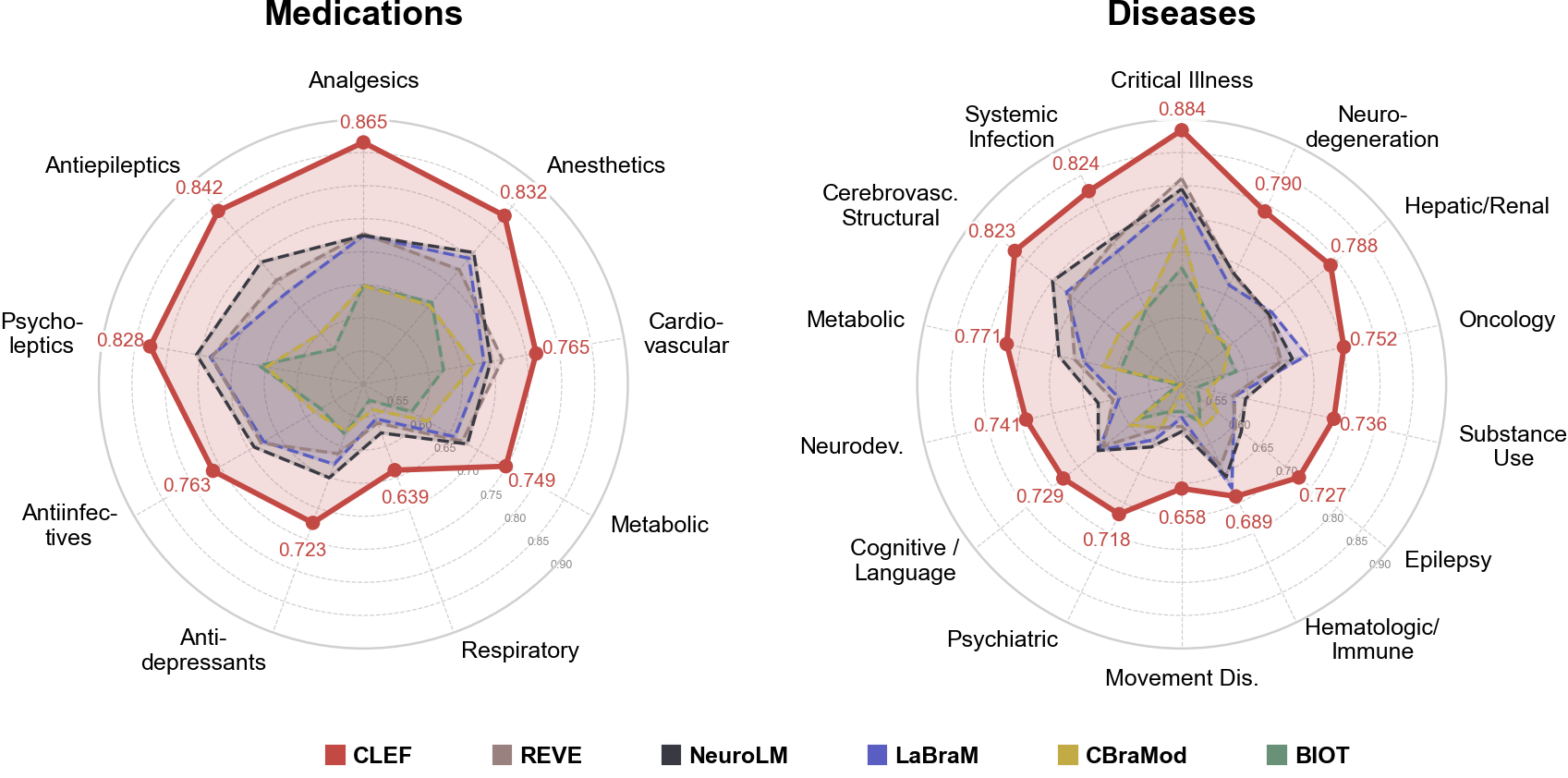}
    \caption{\textbf{Model Performance}. CLEF downstream performance across categories. Probing performance for top 5 downstream-task are averaged within each category. Detailed tasks are listed in Appendix~\ref{sec:top-5}.}
    \label{fig:intro}
    \vspace{-5mm}
\end{figure}

A further challenge is evaluation. Existing EEG benchmarks focus on short-segment BCI tasks, rather than patient-level clinical tasks. We therefore curate a broad clinical EEG benchmark from the Harvard Electroencephalography Database (HEEDB), a large multi-hospital resource containing over 260{,}000 EEG recordings from more than 108{,}000 patients, many linked to deidentified neurophysiology reports and EHR data~\citep{sun2025harvard}. We construct 234 downstream classification tasks spanning three clinical axes: disease phenotypes, medication exposures, and clinically meaningful EEG features. 

Empirically, CLEF establishes a strong foundation model for clinical EEG. Across the 234-task benchmark, CLEF outperforms recent EEG foundation model baselines on 229 of 234 tasks, improving mean AUROC from 0.65 for the strongest baseline to 0.74, as summarized in Fig.~\ref{fig:intro}.   Importantly, CLEF's performance gains generalize to new diseases and medications that do not appear in the reports or EHRs used for training. They also transfer to external datasets unseen during training. Ablations show that reconstruction-only pretraining already yields representations stronger than prior EEG foundation models, while clinical report and EHR alignment provide additional gains, supporting the central hypothesis that clinical EEG foundation models benefit from both session-scale signal modeling and clinical grounding.

Our contributions are:
\vspace{-2mm}
\begin{itemize}
    \item We formulate \emph{clinical EEG foundation modeling} as patient-level representation learning from full EEG sessions, rather than short-window signal decoding.
    \item We introduce CLEF, a clinically-grounded long-context EEG foundation model that uses multi-channel spectrogram tokens and aligns session-level representations with EEG reports and structured EHR data.
    \item We curate a 234-task clinical EEG benchmark spanning disease phenotypes, medication exposures, and EEG findings, enabling systematic evaluation of whether EEG representations capture clinically meaningful information.
    \item We show that CLEF outperforms prior EEG foundation models across the benchmark and generalizes to multiple external datasets, establishing full-session, clinically grounded EEG representation learning as a benchmarkable foundation-model problem.
\end{itemize}

\vspace{-2mm}
\vspace{-2mm}

\section{Related Work}
\vspace{-2mm}

\textbf{EEG Foundation Models.} Foundation-model methods have recently been adapted to EEG through contrastive learning, masked reconstruction, and autoregressive prediction~\cite{kostas2021bendr,chien2022maeeg,cui2024neuro,jiang2024large,chen2024eegformer,wang2025cbramod,wang2024eegpt,zhou2025csbrain,yang2025thdbar,ouahidi2025reve}. Representative models include LaBraM, which learns discrete neural tokens for masked modeling~\cite{jiang2024large}; CBraMod, which models spatial and temporal structure with a criss-cross transformer~\cite{wang2025cbramod}; and REVE, which emphasizes robustness across heterogeneous electrode montages with large-scale pretraining~\cite{ouahidi2025reve}. These models have substantially advanced EEG representation learning, but they are primarily designed for the BCI-style setting: decoding labels from short windows of neural activity, typically a few seconds long. Clinical EEG interpretation poses a different representation-learning problem. Neurologists examine full recording sessions, integrate patterns that recur or evolve over time, and interpret the signal in clinical context. CLEF therefore departs from prior EEG foundation models in both temporal scale and supervision: it learns patient-level representations from full EEG sessions and grounds them in clinical semantics rather than relying only on signal-intrinsic objectives.

\textbf{Multimodal Clinical Representation Learning.} Contrastive multimodal learning has become a powerful approach for learning transferable representations, beginning with image--text alignment in CLIP~\cite{radford2021learning} and extending to medical domains through alignment of clinical images with reports or biomedical text~\cite{zhang2022contrastive,tiu2022expert,zhang2023biomedclip,eslami2023pubmedclip}. In EEG, multimodal learning has mainly focused on perceptual or task-driven settings, such as decoding text from brain activity, aligning EEG with visual-language embeddings, or using language to unify short-segment EEG tasks~\cite{wang2022open,duan2023dewave,ma2025brainclip,jiang2025neurolm}. CLEF instead uses multimodal learning for clinical grounding. Neurologist reports provide expert descriptions of electrographic findings, while EHR data provide complementary information about diagnoses, medications, and demographics. 
Aligning full-session EEG embeddings with these two modalities allows CLEF to learn representations organized around the clinical concepts used in EEG interpretation. To our knowledge, prior EEG foundation models have not combined session-scale modeling with report and EHR alignment for patient-level clinical EEG representation learning.
\vspace{-2mm}

\section{Method}
\label{sec:method}
\begin{figure}
    \centering
    \includegraphics[width=\linewidth]{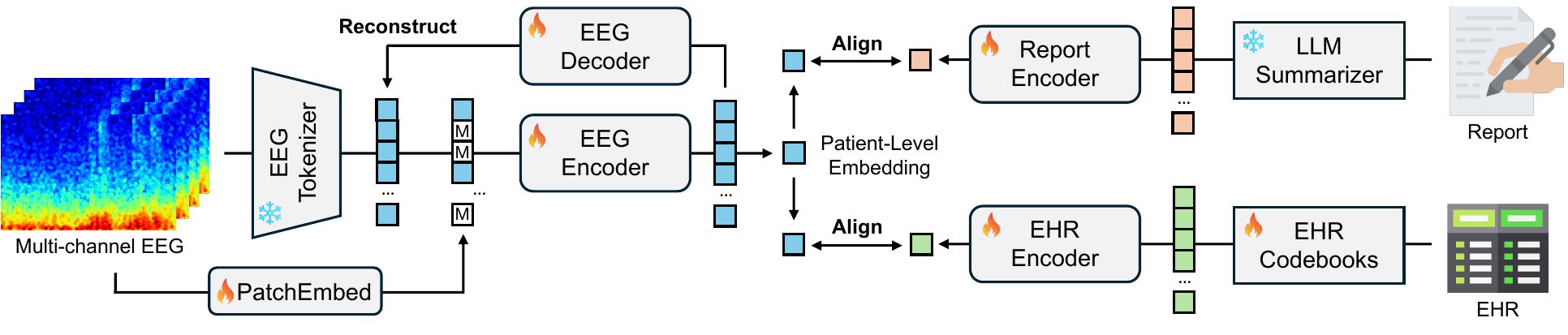}
    \vspace{-5mm}
    \caption{\textbf{CLEF Pipeline.} \raisebox{-0.15ex}{\includegraphics[height=0.9em]{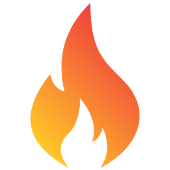}} and \raisebox{-0.15ex}{\includegraphics[height=0.9em]{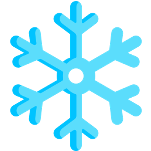}} denote trainable and frozen components, respectively. \textbf{Stage~I} (left) pretrains the EEG encoder by masked reconstruction over a sequence of multi-channel spectro-temporal codes produced by a frozen tokenizer, with a parallel patch-embedding branch reinjecting details lost to quantization. \textbf{Stage~II} (right) grounds the resulting patient-level embedding in clinical semantics by aligning it with two modalities, LLM-summarized neurologist reports and EHR codes, through symmetric contrastive losses.}
    \label{fig:pipeline}
    \vspace{-5mm}
\end{figure}

\vspace{-2mm}
CLEF is built on the two principles: \emph{long-context modeling} at the scale of a full EEG session, and \emph{grounding in clinical semantics}. These map onto a two-stage pipeline (Figure~\ref{fig:pipeline}). Stage~I makes session-scale modeling tractable by learning a 3D compact spectro-temporal token representation through masked reconstruction. Stage~II grounds the resulting patient-level embedding in clinical meaning by aligning it with two clinical modalities: free-text neurologist reports and structured EHR data. The design choices introduced below are validated empirically in Appendix~\ref{sec:ablation}.

\vspace{-2mm}
\subsection{Stage I: Session-Scale Masked Reconstruction}
\label{sec:stage1}
\vspace{-2mm}

A 20-minute session of EEG recording contains nearly five million raw samples. In addition, the instantaneous phase of EEG is largely stochastic, so two recordings from the same patient under identical conditions share spectral content but differ significantly in waveform. Stage~I addresses both issues. We first move into the time-frequency domain via multitaper spectrogram, which discards phase while preserving the spectral features neurologists actually interpret (sleep spindles, slowing, alpha rhythms), and yields more stationary local statistics than the raw waveform. We convert each channel of an EEG recording $\mathbf{X} \in \mathbb{R}^{C \times T}$ to a log-magnitude multitaper spectrogram~\cite{thomson1982spectrum} over the clinically relevant 0--32\,Hz band, yielding $\mathbf{S} \in \mathbb{R}^{C \times H \times W}$ with $C{=}19$ channels, $H{=}128$ frequency bins (0.25\,Hz resolution), and $W{=}2{,}048$ time frames spanning a 1{,}280-second session. Even after this transform, a session-scale spectrogram remains too large to feed directly into a Transformer; Stage~I therefore learns a discrete tokenizer that compresses the multi-channel spectrogram into a short sequence of 3D codes, and pretrains the EEG encoder to predict masked codes. 
\vspace{-2mm}
\subsubsection{Multi-Channel Spectrogram Tokenization}
\label{sec:tokenizer}
\vspace{-1mm}

We train the tokenizer within the VQGAN framework~\cite{esser2021taming}: an encoder $E$ maps $\mathbf{S}$ to continuous latents $E(\mathbf{S}) \in \mathbb{R}^{d \times H' \times W'}$, which are quantized against a learned codebook $\mathcal{V} = \{\mathbf{e}_k\}_{k=1}^{K}$ via $\mathbf{z}_{h,w} = \mathbf{e}_{k^\star}$, $k^\star = \arg\min_{k} \| E(\mathbf{S})_{h,w} - \mathbf{e}_k \|_2$. A decoder $G$ reconstructs $\hat{\mathbf{S}} = G(\mathbf{Z})$, and a PatchGAN discriminator $D$ provides adversarial supervision:
\begin{equation}
\resizebox{0.8\columnwidth}{!}{
    $\mathcal{L}_{\text{tok}} = \mathcal{L}_{\text{rec}}(\mathbf{S}, \hat{\mathbf{S}}) + \lambda_{\text{code}} \|\mathrm{sg}[E(\mathbf{S})] - \mathbf{Z}\|_2^2 + \lambda_{\text{commit}} \|E(\mathbf{S}) - \mathrm{sg}[\mathbf{Z}]\|_2^2 + \lambda_{\text{adv}} \mathcal{L}_{\text{adv}}(D, G, E).$}
    \label{eq:tokloss}
\end{equation}
Three aspects of this framework require adaptation to multi-channel clinical EEG: how channels are arranged at the input, how missing channels are handled, and how the reconstruction loss is designed.

\textbf{Joint multi-channel tokenization.} Tokenizing each of the $C$ channels independently ignores strong spatial dependencies across electrodes and produces a token sequence growing linearly with $C$. We instead stack the $C$ channels as the input channels of a 2D convolutional encoder, analogous to RGB in a natural image: every filter in the first layer sees all electrodes simultaneously, and each discrete code summarizes a time--frequency region \emph{across the entire montage}. This compresses a 20-minute session into $N = H'W' = 2{,}048$ tokens, making session-scale Transformer modeling tractable.

\textbf{Channel masking.} Clinical EEG is not always acquired on the full montage. To expose the tokenizer to missing channels, we apply random channel masking to its input while keeping the reconstruction target and the discriminator's real sample as the full spectrogram, training the tokenizer to \emph{impute} missing channels. The masking probability is ramped linearly from zero to its target value over the first portion of training, since applying full masking from the start forces the model to fight an ill-posed imputation problem before its encoder has learned any useful spatial structure.

\textbf{Inter-channel-emphasized reconstruction loss.} A naive $\ell_1$ loss $\|\mathbf{S} - \hat{\mathbf{S}}\|_1$ treats every (channel, frequency, time) entry equally, which sounds neutral but is in fact heavily biased toward what is \emph{shared} across channels. Each value decomposes into a mean component (across channels) and a differential component (deviations from the mean); a channel-wise $\ell_1$ penalty counts the mean in every one of the $C$ summands but each differential in only one, rewarding the model far more for matching the across-channel average than for reproducing the inter-channel structure that defines many clinical phenomena (e.g., focal discharges, lateralization, topographic gradients). We therefore decompose the reconstruction target explicitly. Let $\bar{\mathbf{S}} = \tfrac{1}{C}\sum_{c} \mathbf{S}^{(c)}$ and $\Delta\mathbf{S}^{(c)} = \mathbf{S}^{(c)} - \bar{\mathbf{S}}$, with $\bar{\hat{\mathbf{S}}}$ and $\Delta\hat{\mathbf{S}}^{(c)}$ defined analogously. The reconstruction loss is
\begin{equation}
\resizebox{0.53\columnwidth}{!}{
    $\mathcal{L}_{\text{rec}}(\mathbf{S}, \hat{\mathbf{S}}) = \bigl\|\bar{\mathbf{S}} - \bar{\hat{\mathbf{S}}}\bigr\|_1 + \frac{\gamma_{\text{diff}}}{C} \sum_{c=1}^{C} \bigl\|\Delta\mathbf{S}^{(c)} - \Delta\hat{\mathbf{S}}^{(c)}\bigr\|_1,$}
    \label{eq:rec}
\end{equation}
with $\gamma_{\text{diff}} > 1$ upweighting the differential term to prevent collapse of inter-channel variation.

\subsubsection{Masked Image Modeling}
\label{sec:mim}
\vspace{-1.5mm}
The pre-trained tokenizer maps each spectrogram to a sequence of codes $(z_1, \dots, z_N)$. Each token $z_i$ at grid position $(h_i, w_i)$ is embedded through a learnable lookup table $\mathbf{W}_{\text{tok}} \in \mathbb{R}^{K \times d}$ and combined with factorized frequency and time positional embeddings $\mathbf{p}^{\text{freq}}_{h_i}, \mathbf{p}^{\text{time}}_{w_i}$. 
To reinject details lost to quantization for Stage~II's cross-modal alignment to exploit, we additionally feed the raw spectrogram patch $\mathbf{s}_i \in \mathbb{R}^{C \times p_h \times p_w}$ at the same grid location through a patch-embedding projection $\mathbf{W}_{\text{patch}}$:
\begin{equation}
\resizebox{0.45\columnwidth}{!}{$
    \mathbf{x}_i = \mathbf{W}_{\text{tok}}[z_i] + \mathbf{W}_{\text{patch}}\,\mathrm{vec}(\mathbf{s}_i) + \mathbf{p}^{\text{freq}}_{h_i} + \mathbf{p}^{\text{time}}_{w_i}.$}
    \label{eq:mim-input}
\end{equation}
We adopt a Transformer encoder--decoder for masked image modeling. Let $\mathcal{M} \subset \{1, \dots, N\}$ denote the set of masked positions, with the masking ratio sampled from a truncated Gaussian $\mathcal{TN}(\mu, \sigma^2)$. At masked positions, $\mathbf{x}_i$ is replaced with a learnable \texttt{[MASK]} token (positional embeddings retained). The encoder $f_{\theta}$ produces contextualized representations $\mathbf{h}_i$, which are pooled to form the patient-level embedding $\mathbf{u}$ and passed to a lightweight Transformer decoder $g_{\phi}$ that predicts masked tokens against the codebook. The MIM objective is the cross-entropy over masked positions:
\begin{equation}
\resizebox{0.75\columnwidth}{!}{
    $\mathcal{L}_{\text{Recon}} = -\frac{1}{|\mathcal{M}|} \sum_{i \in \mathcal{M}} \log \hat{p}_{i}[z_i], \quad \hat{p}_i = \mathrm{softmax}\bigl(\mathrm{Head}(g_{\phi}(\mathbf{h})_i)\, \mathbf{W}_{\text{tok}}^\top + \mathbf{b}\bigr).$}
    \label{eq:mim-loss}
\end{equation}
After Stage~I, $g_{\phi}$ is discarded and only the encoder $f_{\theta}$ is carried into Stage~II.

\vspace{-1mm}
\subsection{Stage II: Cross-Modal Alignment with Report and EHR}
\label{sec:stage2}
\vspace{-2mm}
Stage~I gives us a session-scale encoder, but signal-only pretraining cannot recover clinical semantics that are not directly visible in the EEG. For example, whether a patient carries a diagnosis of Alzheimer's disease leaves signatures in the recording, but learning to read them requires supervision from the modalities that encode this information. Stage~II therefore aligns the patient-level embedding $\mathbf{u} \in \mathbb{R}^{d}$ with two complementary clinical modalities: \emph{free-text neurologist reports}, which capture expert interpretation of the recording itself, and \emph{structured EHR data}, which describe the patient's broader clinical state. We project $\mathbf{u}$ through two linear heads $\pi_{\text{rep}}, \pi_{\text{ehr}} : \mathbb{R}^{d} \to \mathbb{R}^{d'}$ into modality-specific subspaces, and enforce alignment in each subspace via a contrastive loss~\cite{radford2021learning}.
\vspace{-2mm}
\subsubsection{Alignment with Free-Text Reports}
\vspace{-1mm}
Raw EEG reports are not directly ingestible by a text encoder: a single report may span several thousand tokens and interleave semi-templated sections (patient history, recording procedure and parameters) with free prose (clinical impression, recommendations), much of it irrelevant to the electrographic content. Direct encoding yields embeddings dominated by boilerplate, and reports' length makes them impractical for contrastive alignment with standard text encoders. We therefore preprocess them with an LLM summarizer, which raises a practical question: among the many plausible combinations of prompt and output length, which best preserves the clinical content?

We answer this question with an automated selection pipeline based on a \emph{QA-consistency score} (details in Appendix \ref{sec:llm}). Let $\mathcal{R} = \{r_1, \dots, r_N\}$ be a held-out pool of raw reports and $\mathcal{Q} = \{q_1, \dots, q_M\}$ a predefined set of questions covering the electrographic content of the recording. For each candidate $(p, \ell)$ specifying a summarization prompt $p$ and output token length $\ell$, we use an LLM $\Phi$ to produce a processed version $\tilde{r}_i = \Phi_{\text{sum}}(r_i; p, \ell)$ for every report, and then parse both raw and processed reports in question-answering mode $\Phi_{\text{qa}}(\cdot, q)$. Per-question agreement $\sigma_{i,j}(p, \ell) = \text{Agree}\bigl(\Phi_{\text{qa}}(r_i, q_j),\, \Phi_{\text{qa}}(\tilde{r}_i, q_j)\bigr)$ is computed using exact match for boolean and integer questions and LLM-judged similarity for free-text questions. We define QA-consistency score as the average of per-question agreement $S(p, \ell) = \tfrac{1}{NM} \sum_{i,j} \sigma_{i,j}(p, \ell)$, and select $(p^\star, \ell^\star) = \arg\max_{(p, \ell)} S(p, \ell)$ to use for the LLM summarizer. Crucially, because $\Phi$ is never exposed to the question set during processing, high consistency cannot be achieved by selectively preserving content that answers these specific questions; performance on this probe should generalize to the broader electrographic content.

Given a summarized report $\tilde{r}$, we obtain its embedding via a frozen T5-base encoder~\cite{raffel2020exploring} followed by learnable self-attention layers and mean pooling: $\mathbf{v}_{\text{rep}} = \text{Pool}\bigl(f_{\text{rep}}(f_{\text{T5}}(\tilde{r}))\bigr) \in \mathbb{R}^{d}$. Alignment is enforced by a symmetric contrastive loss~\cite{radford2021learning} between $\pi_{\text{rep}}(\mathbf{u})$ and $\mathbf{v}_{\text{rep}}$:
\begin{equation}
\resizebox{0.8\columnwidth}{!}{$
    \mathcal{L}_{\text{Report}} = -\frac{1}{2B}\sum_{i=1}^{B} \left[ \log \frac{\exp\bigl(\langle \pi_{\text{rep}}(\mathbf{u}_i), \mathbf{v}_{\text{rep},i}\rangle / \tau \bigr)}{\sum_{j} \exp\bigl(\langle \pi_{\text{rep}}(\mathbf{u}_i), \mathbf{v}_{\text{rep},j}\rangle / \tau\bigr)} + \log \frac{\exp\bigl(\langle \pi_{\text{rep}}(\mathbf{u}_i), \mathbf{v}_{\text{rep},i}\rangle / \tau \bigr)}{\sum_{j} \exp\bigl(\langle \pi_{\text{rep}}(\mathbf{u}_j), \mathbf{v}_{\text{rep},i}\rangle / \tau\bigr)} \right].
$}
    \label{eq:report-clip}
\end{equation}

\subsubsection{Alignment with Structured EHR}
\vspace{-1mm}
The Electronic Health Record (EHR) provides a complementary view to the report: rather than interpreting the current recording, it describes the patient's overall clinical profile through tabular entries for demographics, medications, and diagnoses. How should such data be presented to a encoder? A natural attempt is to serialize the record into a sentence (e.g., ``65-year-old female, on levetiracetam and propofol, with a history of epilepsy and chronic kidney disease\dots'') and use a text encoder. However, tabular EHRs are unordered sets of codes from controlled vocabularies. Serialization imposes an arbitrary order that the transformer would treat as informative, and splits each code into subword pieces the encoder must learn to regroup, wasting capacity and compute.

We instead encode the EHR according to its native structure. Each patient is a set of discrete entries from three sources: a \emph{demographic} group assignment (age bucket, sex, race), a set of \emph{medication codes}, and a set of \emph{diagnosis codes}. We maintain learnable codebooks $\mathbf{E}_{\text{med}} \in \mathbb{R}^{K_{\text{med}} \times d}$ and $\mathbf{E}_{\text{dx}} \in \mathbb{R}^{K_{\text{dx}} \times d}$ with one entry per distinct code, plus demographic codebooks $\mathbf{E}_{\text{age}}, \mathbf{E}_{\text{sex}}, \mathbf{E}_{\text{race}}$ with one entry per group. This design represents each code by \emph{identity} rather than subwords, learns embeddings end-to-end against the EEG signal so clinically related codes can organize themselves, and is permutation-invariant by construction, matching the set semantics of the data.

For a patient with demographic indices $(a, s, r)$, active medications $\mathcal{M}$, and diagnoses $\mathcal{D}$, we concatenate the looked-up embeddings into a variable-length sequence
\begin{equation}
\mathbf{X}_{\text{ehr}} = \big[\,\mathbf{E}_{\text{age}}[a],\, \mathbf{E}_{\text{sex}}[s],\, \mathbf{E}_{\text{race}}[r],\; \{\mathbf{E}_{\text{med}}[m]\}_{m \in \mathcal{M}},\; \{\mathbf{E}_{\text{dx}}[d]\}_{d \in \mathcal{D}}\,\big],
\end{equation}
which is processed by self-attention layers and mean-pooled to yield $\mathbf{v}_{\text{ehr}} = \text{Pool}(f_{\text{ehr}}(\mathbf{X}_{\text{ehr}})) \in \mathbb{R}^{d}$. Self-attention lets codes \emph{contextualize} one another; a diagnosis of epilepsy, for instance, carries different implications with vs.\ without antiepileptic medications.

Alignment is enforced by a second symmetric contrastive loss $\mathcal{L}_{\text{EHR}}$ between $\pi_{\text{ehr}}(\mathbf{u})$ and $\mathbf{v}_{\text{ehr}}$, defined analogously to Equation~\ref{eq:report-clip}. The total Stage~II objective is
\begin{equation}
    \mathcal{L}_{\text{Align}} = \mathcal{L}_{\text{Report}} + \mathcal{L}_{\text{EHR}}.
    \label{eq:stage2}
\end{equation}
\subsubsection{Training Scheme}
\vspace{-1mm}
Stages~I and~II are trained sequentially: we first train the encoder with $\mathcal{L}_{\text{Recon}}$ to convergence, then continue training it with $\mathcal{L}_{\text{Align}}$, because the two objectives operate at different levels of abstraction: Masked reconstruction is a dense, position-level objective that forces the encoder to capture the intrinsic spectro-temporal structure of the recording, whereas the contrastive losses provide a sparse, sequence-level signal that is more informative when such structure exists. Throughout both stages, we randomly drop a fraction of tokens from the encoder input for regularization and efficiency~\cite{li2023scaling}, and apply random channel masking to expose the encoder to partial-montage conditions.
\vspace{-2mm}
\section{Experiments}
\label{sec:exp}
\vspace{-2mm}

We evaluate whether CLEF's learned representations capture
clinically meaningful information by probing on a broad suite
of downstream tasks. Throughout, the encoder is frozen after
pretraining and a lightweight probing head is trained on top
of the patient-level embeddings for each task, isolating the quality of
the learned representation from task-specific finetuning. We compare
CLEF against state-of-the-art EEG foundation models under identical
probing and aggregation protocols. Implementation details of model and training are provided in Appendix~\ref{sec:model-training-details}.
\vspace{-2mm}
\subsection{Datasets and Benchmark}
\label{sec:datasets}
\vspace{-2mm}
\textbf{Pretraining.} We pretrain CLEF on a snapshot of the Harvard
Electroencephalography Database (HEEDB)~\cite{sun2025harvard}, comprising
260{,}604 recordings totaling 2M hours from 108{,}341 patients with EHR data, of which 59.3\% have
linked neurologist reports. Patients are split 80/10/10 into
train/validation/test, with all sessions from the same patient kept
in the same split. The same split is used for both pretraining and downstream benchmark evaluation, so no test patient is seen during pretraining.

\textbf{Benchmark.} We construct 234 patient-level binary classification tasks from HEEDB
spanning three axes of clinical meaning: (i) 120 disease phenotypes tasks derived from ICD-10 codes grouped into phecodes~\cite{denny2013systematic}, (ii) 99 medication exposures tasks derived from prescription records mapped to ATC
codes~\cite{whocc_atc}, and (iii) 15 EEG feature tasks
extracted from neurologist reports. For each task, each patient contributes at most
one session, and up to 10 controls per positive case are \textit{matched} on
age, sex, site, and recording setting (ICU, EMU, Routine EEG) to reduce confounding.
Full procedures for benchmark construction are in
Appendix~\ref{sec:benchmark-processing}.

\textbf{Held-out clinical concepts.}  To assess transferability to new diseases, medications, and EEG features unseen during training, we run an additional pretraining experiment in which 20\% of the tasks in the benchmark are randomly selected and the relevant terms are removed from both reports and EHRs. This
yields a held-out set of clinical concepts that the model has never been
exposed to during alignment, allowing us to evaluate transfer to
novel concepts.

\textbf{External cohorts.} To assess generalization across data sources and montages, we evaluate on three external cohorts: TUAB (abnormal EEG detection) and TUEP (epilepsy detection) from the TUH EEG Corpus~\cite{obeid2016temple}, both full-montage; and HSP from the PhysioNet Challenge 2026~\cite{hsp}, a 6-channel cognitive impairment detection dataset. Each dataset was split at patient level into 75\% training and 25\% test, with 15\% of the training data held out for validation, and evaluated at patient level.

\vspace{-2mm}
\subsection{Evaluation Protocol}
\label{sec:protocol}
\vspace{-2mm}
\textbf{Baselines.} 
We compare CLEF against five state-of-the-art EEG foundation models that collectively represent the leading approaches to large-scale EEG pretraining: BIOT~\cite{yang2023biot}, a biosignal transformer trained on heterogeneous channel configurations; LaBraM~\cite{jiang2024large}, a large brain model pretrained via masked neural code prediction; CBraMod~\cite{wang2025cbramod}, which introduces criss-cross attention to capture spatial-temporal dependencies; REVE~\cite{ouahidi2025reve}, a recent encoder designed for robust EEG representation learning; and NeuroLM~\cite{neurolm}, which unifies EEG and language modeling within a shared architecture. For each baseline, we probe on top of the publicly released pretrained encoder. NeuroLM's instruction-tuned checkpoint is not released, so we reproduced it using the official code. 

\textbf{Probing.} For each task, we train a 2-layer MLP head on the frozen patient-level embedding. To ensure that all methods see the same temporal context regardless of their native training window, every encoder is applied in a sliding-window fashion over the full session and the resulting segment embeddings are mean-pooled into a single patient-level embedding before being passed to the probe. 

\textbf{Metrics.} We report AUROC across all benchmark tasks. For
external datasets, where label prevalence is not controlled, we
also report balanced accuracy~(BACC), with the decision threshold
selected on validation set. Probing experiments are repeated
across 4 seeds to report mean and standard deviation.

\vspace{-2mm}
\subsection{Benchmark Results}
\begin{figure}[t]
\centering

\begin{subfigure}{1.0\textwidth}
    \centering
    \includegraphics[width=\linewidth]{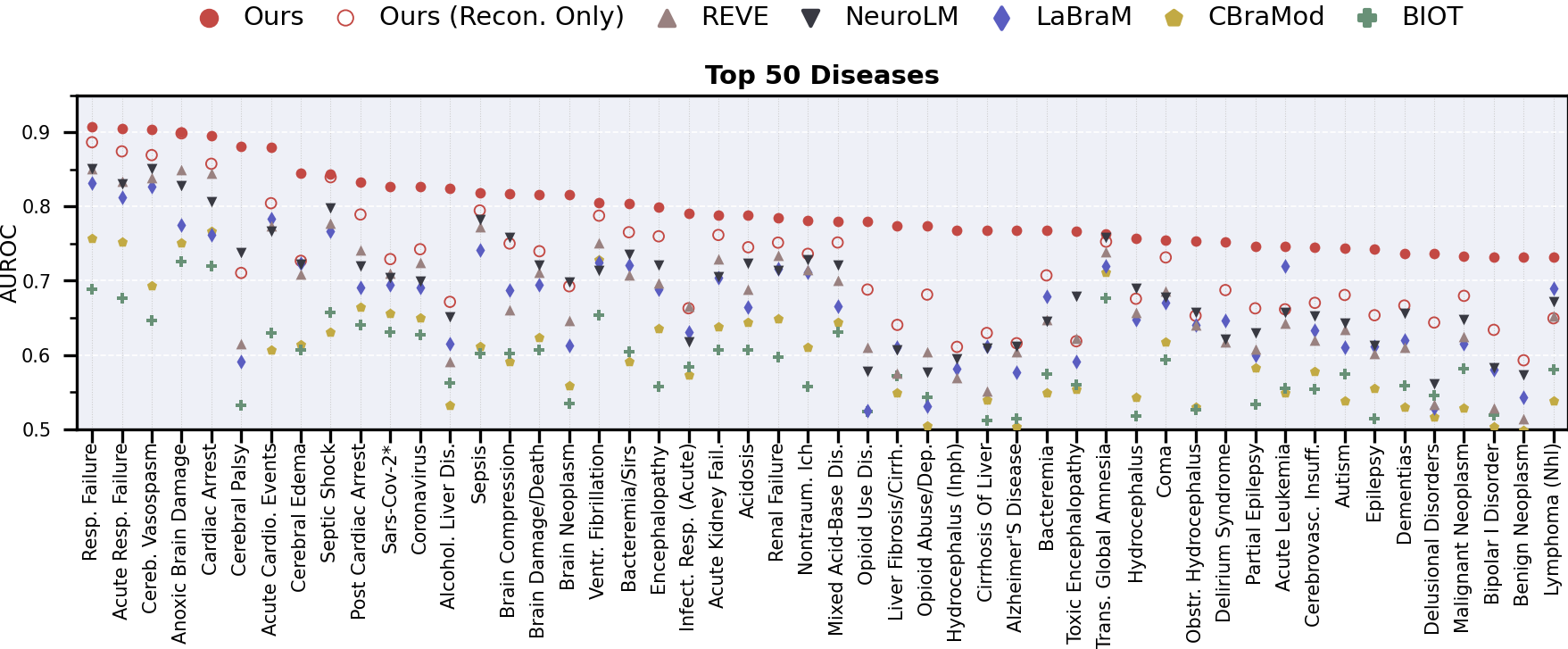}
    \label{fig:panel-a}
\end{subfigure}

\vspace{-1.0em}

\begin{subfigure}{1.0\textwidth}
    \centering
    \includegraphics[width=\linewidth]{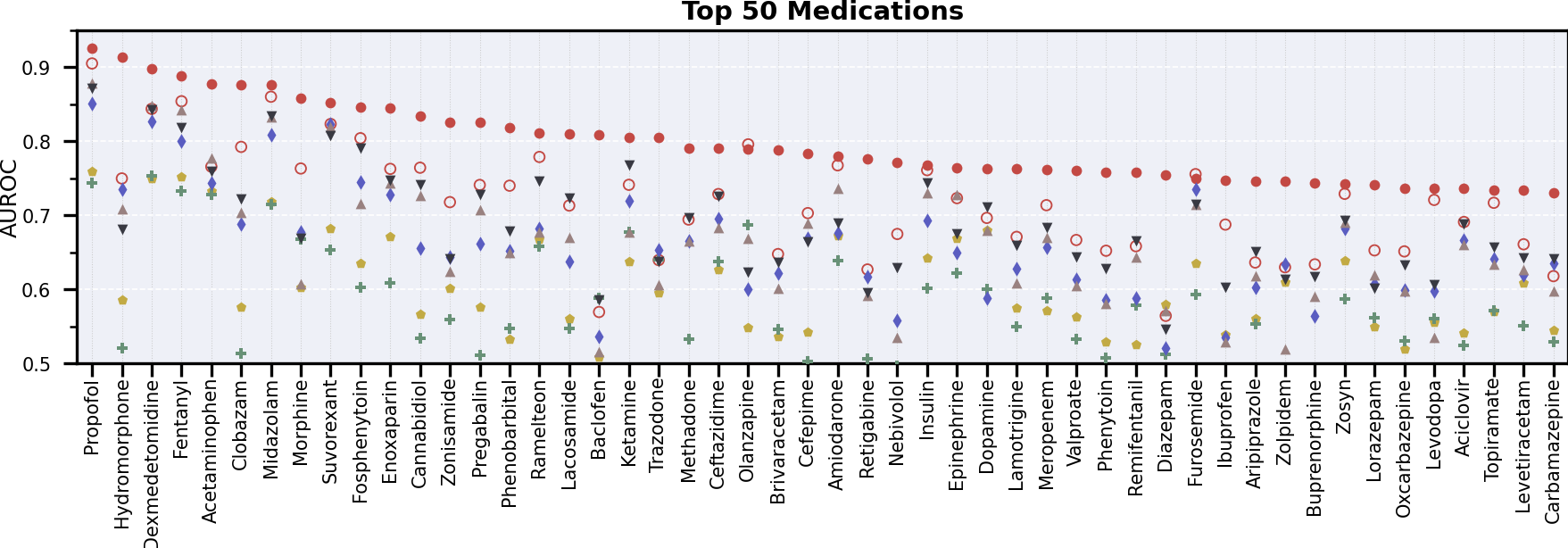}
    \label{fig:panel-b}
\end{subfigure}

\vspace{-1.0em}

\hspace{-0.4cm}
\begin{subfigure}[t]{0.3\textwidth}
    \centering
    \includegraphics[width=\linewidth]{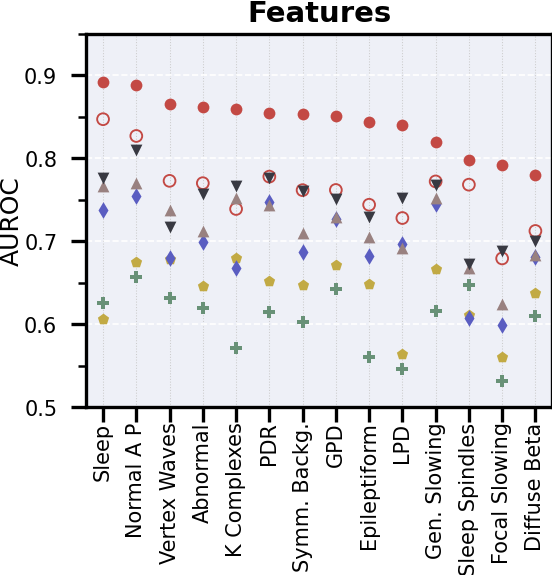}
    \label{fig:panel-c}
\end{subfigure}
\hspace{3mm}
\begin{subfigure}[t]{0.68\textwidth}
  \centering
  \vspace{-4.1cm} 
  \renewcommand{\arraystretch}{1.3}
  \footnotesize
  \setlength{\tabcolsep}{4pt} 
  \resizebox{\linewidth}{!}{%
  \begin{tabular}{l c  c c c}
      \toprule                                                                                                                                                                 
      \textbf{Method} &                                                                                                                                                        
      \textbf{Features} &                                                                                                                                                      
      \textbf{Diseases} &
      \textbf{Medications} &
      \textbf{Overall} \\    
      \midrule 
      BIOT~\cite{yang2023biot}    & 0.603\stdv{0.022} & 0.554\stdv{0.027} & 0.557\stdv{0.030} & 0.559\stdv{0.028} \\                                                                                           
      CBraMod~\cite{wang2025cbramod} & 0.633\stdv{0.007} & 0.559\stdv{0.016} & 0.562\stdv{0.021} & 0.565\stdv{0.017} \\                                                                                                                                                              
      LaBraM~\cite{jiang2024large}  & 0.686\stdv{0.010} & 0.626\stdv{0.012} & 0.627\stdv{0.016} & 0.630\stdv{0.014} \\                                                                                                                                                                
      REVE~\cite{ouahidi2025reve}    & 0.712\stdv{0.011} & 0.622\stdv{0.016} & 0.622\stdv{0.019} & 0.628\stdv{0.017} \\    
      NeuroLM~\cite{neurolm}    & 0.737\stdv{0.006} & 0.635\stdv{0.012} & 0.645\stdv{0.013} & 0.646\stdv{0.012} \\    
      \midrule
      CLEF~(Recon Only) & 0.754\stdv{0.005} & 0.662\stdv{0.007} & 0.669\stdv{0.009} & 0.671\stdv{0.008} \\                                                                                                                                                                                                                                 
      CLEF~(Recon$\to$Align)    & \textbf{0.833}\stdv{0.003} & \textbf{0.728}\stdv{0.006} & \textbf{0.735}\stdv{0.007} & \textbf{0.737}\stdv{0.006} \\                                                                               
      \bottomrule
  \end{tabular} }
\end{subfigure}

\vspace{-1.0em}
\caption{\textbf{Task probing performance across the benchmark.} AUROC for CLEF (full and reconstruction-only variants) and five EEG foundation model baselines on (a) top 50 disease tasks, (b) top 50 medication tasks (c) all EEG feature tasks, sorted by CLEF performance. (d) AUROC across tasks, within each category and overall.} 
\label{fig:internal}
\vspace{-5mm}
\end{figure}

\vspace{-2mm}
\label{sec:main-results}
  Figure \ref{fig:internal} summarizes CLEF probing performance across diseases, medications, and features, and compares it with the baselines under the same probing protocol. Specifically, we show the AUROC for the top 50 diseases, top 50 medications, and all features, and summarize the AUROC across all 234 tasks. Appendix~\ref{sec:detailed-results} details the AUROC for each task. CLEF significantly outperforms the baselines across all categories. It is particularly effective at detecting critical illnesses, systemic infections, and cerebrovascular and cerebrostructural diseases, as well as exposure to medications including analgesics, anesthetics, anti-epileptics, and psycholeptics. Large performance gain across diseases include Cerebral Palsy (+0.27), Alcoholic Liver Disease (+0.21), Alzheimer’s Disease (+0.16), Cirrhosis (+0.16), Cerebral Edema (+0.12). Performance gain across medications include Morphine (+0.18), Zonisamide (+0.18), Hydromorphone (+0.18), Clobazam (+0.17).

Figure~\ref{fig:internal} also shows that CLEF with reconstruction only (i.e., Stage I alone) outperforms the baselines. This indicates that the long-context design and Stage I training are valuable even without cross-modal alignment. To investigate this point further and since the baselines and CLEF each is trained on different datasets, we retrain REVE, the current SOTA for EEG foundation models~\cite{ouahidi2025reve}, on the same dataset used for training CLEF. The AUROC improves from  $0.628 \pm 0.017$ to $0.655 \pm 0.014$, indicating a benefit for the larger and more diverse training dataset in our benchmark. However, the AUROC stays significantly lower than both CLEF with reconstruction only at $0.671 \pm 0.008$, and CLEF with both reconstruction and cross-modal alignment at $0.737 \pm 0.006$.

Notably, some tasks perform well with reconstruction alone and gain little from alignment to clinical semantics in reports and EHRs: propofol reaches an AUROC of 0.905 and anoxic brain injury 0.899, with small improvements after cross-modal alignment to 0.925 and 0.901, respectively. These labels may correspond to visually salient EEG patterns, allowing masked reconstruction to learn features that already support strong classification. In contrast, cerebral palsy improves from 0.711 to 0.881 and trazodone from 0.640 to 0.806 after cross-modal alignment. These larger gains suggest that report and EHR alignment is particularly useful for labels whose EEG associations are not immediately clear from the EEG signal. This pattern is consistent with clinical practice: EEG is commonly used to monitor propofol effects and assess anoxic brain injury, but not for diagnosing cerebral palsy or identifying trazodone exposure~\cite{coma, propofol, trazodone, cerebralpalsy}.

\subsection{Transfer to held-out clinical concepts}
\vspace{-2mm}
\label{sec:held-out-tasks}
\begin{wraptable}{r}{0.50\linewidth}
  \vspace{-1.5cm}
  \renewcommand{\arraystretch}{1.05}
  \centering
  \footnotesize
  \setlength{\tabcolsep}{3pt}
  \caption{
  \textbf{Held-out clinical concept performance.} AUROC on tasks excluded from cross-modal alignment.
  }
  \vspace{-1mm}
  \resizebox{\linewidth}{!}{
  \begin{tabular}{lccc}
    \toprule
    & \textbf{Diseases} & \textbf{Medications} & \textbf{Features} \\
    & ($N=20$) & ($N=21$) & ($N=4$) \\
    \midrule
    BIOT~\cite{yang2023biot} & 0.576\stdv{0.026} & 0.563\stdv{0.032} & 0.624\stdv{0.026} \\
    CBraMod~\cite{wang2025cbramod} & 0.587\stdv{0.013} & 0.582\stdv{0.026} & 0.646\stdv{0.003} \\
    LaBraM~\cite{jiang2024large} & 0.663\stdv{0.008} & 0.640\stdv{0.013} & 0.723\stdv{0.003} \\
    REVE~\cite{ouahidi2025reve} & 0.655\stdv{0.012} & 0.624\stdv{0.020} & 0.733\stdv{0.007} \\
    NeuroLM~\cite{neurolm} & 0.665\stdv{0.007} & 0.652\stdv{0.012} & 0.749\stdv{0.004} \\
    \midrule
    CLEF (No Holdout) & 0.737\stdv{0.004} & 0.733\stdv{0.007} & 0.836\stdv{0.002} \\
    CLEF (Holdout) & 0.739\stdv{0.005} & 0.738\stdv{0.006} & 0.840\stdv{0.001} \\
    \bottomrule
  \end{tabular}
  }
  \label{tab:holdout_comparison}
  \vspace{-1em}
\end{wraptable}

Foundation models should support downstream tasks that are not enumerated during pretraining. 
We therefore evaluate CLEF under a held-out concept protocol. We randomly select 20\% of the 
benchmark tasks and remove their corresponding clinical concepts from cross-modal alignment 
before retraining the model: matching EHR codes are excluded from the structured EHR input, and 
matching report-derived terms are scrubbed from the report targets. The held-out tasks are not 
used during representation learning and are introduced only at probing time, when frozen 
patient-level embeddings are probed using the same protocol as in Section~\ref{sec:protocol}.

Table~\ref{tab:holdout_comparison} shows that CLEF transfers to held-out concepts with no degradation across the 45 held-out tasks.
This indicates that CLEF does not rely on task-specific exposure during alignment to get strong downstream performance and can generalize to novel clinical concepts, suggesting that it learns a broadly transferable representation of clinical EEG.

\subsection{Generalization to External Cohorts}
\vspace{-2mm}
\label{sec:external-cohorts}

To further access whether CLEF's learned representation generalize to recordings collected at other institutions and with different montages, we evaluate CLEF on three external cohorts, each labeled for a different task.
Note that TUAB and TUEP are derived from the TUH corpus, which is included in the pretraining data of all baselines, giving them an in-distribution advantage on these tasks. 
Table~\ref{tab:holdout_main_results} shows CLEF achieves the best AUROC and BACC on all three datasets, improving
over the strongest baseline by +3.1/+2.6 points on TUAB, +5.6/+6.4 on TUEP,
and +8.2/+8.5 on HSP (AUROC/BACC), suggesting CLEF learns representations generalize across institutions and montages.
\begin{table}[t]
  \centering
  \footnotesize
  \caption{\textbf{Probing Performance on external cohorts.} Best results per column are in \textbf{bold}.}
  \vspace{1mm}
  \label{tab:holdout_main_results}
  \setlength{\tabcolsep}{6pt}
  \resizebox{0.85\textwidth}{!}{%
    \begin{tabular}{l cc cc cc}
    \toprule
    & \multicolumn{2}{c}{\textbf{TUAB}} & \multicolumn{2}{c}{\textbf{TUEP}} & \multicolumn{2}{c}{\textbf{HSP}} \\
    & \multicolumn{2}{c}{Abnormal EEG} & \multicolumn{2}{c}{Epilepsy} & \multicolumn{2}{c}{Cognitive Impairment} \\
    \cmidrule(lr){2-3} \cmidrule(lr){4-5} \cmidrule(lr){6-7}
    \textbf{Method} & AUROC$^\uparrow$ & BACC$^\uparrow$ & AUROC$^\uparrow$ & BACC$^\uparrow$ & AUROC$^\uparrow$ & BACC$^\uparrow$ \\
    \midrule
    BIOT~\cite{yang2023biot}    & 0.853\stdv{0.003} & 0.786\stdv{0.006} & 0.821\stdv{0.019} & 0.731\stdv{0.021} & 0.607\stdv{0.028} & 0.571\stdv{0.024} \\
    LaBraM~\cite{jiang2024large}  & 0.876\stdv{0.001} & 0.815\stdv{0.005} & 0.855\stdv{0.002} & 0.780\stdv{0.003} & 0.567\stdv{0.068} & 0.549\stdv{0.050} \\
    CBraMod~\cite{wang2025cbramod} & 0.840\stdv{0.000} & 0.719\stdv{0.025} & 0.855\stdv{0.005} & 0.763\stdv{0.003} & 0.578\stdv{0.025} & 0.562\stdv{0.038} \\
    REVE~\cite{ouahidi2025reve}    & 0.904\stdv{0.004} & 0.811\stdv{0.015} & 0.904\stdv{0.011} & 0.765\stdv{0.019} & 0.597\stdv{0.037} & 0.567\stdv{0.012} \\
    NeuroLM~\cite{neurolm}    & 0.908\stdv{0.001} & 0.825\stdv{0.002} & 0.903\stdv{0.005} & 0.772\stdv{0.013} & 0.586\stdv{0.017} & 0.550\stdv{0.011} \\
    CLEF~(Ours)    & \textbf{0.939}\stdv{0.001} & \textbf{0.851}\stdv{0.002} & \textbf{0.960}\stdv{0.000} & \textbf{0.844}\stdv{0.006} & \textbf{0.689}\stdv{0.012} & \textbf{0.656}\stdv{0.007} \\
    \bottomrule
  \end{tabular}%
  }
  \vspace{-4mm}
\end{table}

\begin{wrapfigure}{r}{0.22\textwidth}
\vspace{-4mm}
\centering
\footnotesize

\captionof{table}{\textbf{Model Scaling.}}
\vspace{-2mm}
\label{tab:model-size}
\begin{tabular}{lc}
\toprule
Model Size & AUROC \\
\midrule
17M   & 0.729 \\
42M   & 0.735 \\
148M  & 0.749 \\
\bottomrule
\end{tabular}

\vspace{1mm}

\captionof{table}{\textbf{Data Scaling.}}
\label{tab:data-size}
\begin{tabular}{lc}
\toprule
Data Size & AUROC \\
\midrule
10\%   & 0.697 \\
25\%   & 0.706 \\
50\%   & 0.720 \\
100\%  & 0.729 \\
\bottomrule
\end{tabular}
\vspace{-8mm}
\end{wrapfigure}
\subsection{Scaling Analysis}
\vspace{-2mm}
We analyze how CLEF scales along both model and data size, reporting mean AUROC over the 234 benchmark tasks with the same seed.

\textbf{Model Size.} We scale the encoder along depth ($L$) and width ($D$): 17M ($L{=}4$, $D{=}512$), 42M (default, $L{=}12$, $D{=}512$), and 148M ($L{=}20$, $D{=}768$). Table~\ref{tab:model-size} shows AUROC rising monotonically with parameter count, with no sign of saturation at the scales explored.

\textbf{Data Size.} Fixing the 17M encoder and the number of training iterations, we vary the fraction of the pre-training corpus from 10\% to 100\%. Table~\ref{tab:data-size} shows a similar monotonic trend, with AUROC rising from 0.697 to 0.729. 

Together, these results indicate that CLEF is currently limited by both model and data scale, and that scaling either axis is a promising direction for further gains. Further robustness analyses for channel count and recording length are provided in Appendix~\ref{sec:analysis}.

\section{Conclusion}
We presented CLEF, a clinically-grounded long-context EEG foundation model that learns patient-level representations by combining session-scale masked image modeling over multitaper spectrogram tokens with cross-modal alignment to neurologist reports and structured EHR data. Trained on over 100k patients and evaluated on a 234-task clinical benchmark, CLEF outperforms prior EEG foundation models on 229 of 234 tasks, and generalize to unseen concepts and external cohorts. These results establish full-session, clinically grounded representation learning as a viable foundation-model paradigm for clinical EEG.
Limitations include lack of real-time support, insufficient time-window for learning sleep architecture, and limited interpretability; we discuss these and additional considerations, along with broader societal impacts, in Appendix~\ref{sec:discussion}.
We hope CLEF draws attention to the diagnostic value locked within full-length EEG recordings and encourages further development in this direction.

{
\small
\bibliographystyle{unsrtnat}
\bibliography{ref}
}
\clearpage
\appendix
\section{Discussion}
\label{sec:discussion}
\subsection{Limitations}
\label{sec:limitations}
\textbf{Lack of Real-time Support.} CLEF summarizes an entire EEG session as a single patient-level vector. In clinical settings, EEG is often used to monitor real-time events such as sedation during surgery. The CLEF framework is not intended for such real-time scenarios and should not be used as a real-time monitoring system.

\textbf{Context Window.}
While the model is trained on 20-minute context windows that are significantly larger than those of prior work, clinical EEG sessions may require even larger context windows to capture longer-timescale dynamics such as full sleep architecture or multi-day cEEG trends.

\textbf{Interpretability.} CLEF's embeddings are not directly interpretable. We do not provide attribution mechanisms linking specific time segments, channels, or frequency bands to embedding dimensions or downstream predictions, which may limit clinical adoption in settings that require explanation of specific findings. Developing such attribution mechanisms is an important direction for future work, particularly for clinical adoption.

\subsection{Broader Impacts}
\label{sec:broader-impacts}
\textbf{Positive Societal Impacts.} Clinical EEG foundation models that produce patient-level representations have the potential to substantially expand access to expert-level EEG interpretation. CLEF could serve as a decision-support layer in under-resourced health systems where neurophysiology expertise is scarce, helping to democratize access to specialist-level assessment. Beyond direct interpretation support, patient-level EEG representations grounded in clinical semantics could accelerate biomarker discovery for conditions whose EEG correlates are suspected but poorly characterized at scale, and enable treatment-response prediction across neurological and psychiatric disorders, helping clinicians personalize therapy and reducing the trial-and-error burden borne by patients. 

\textbf{Negative Societal Impacts.}
Deploying a clinical EEG foundation model also carries meaningful risks. 
Because CLEF's representations are aligned with clinical reports and EHR data, they can encode and propagate distributional biases tied to local patient demographics, recording equipment, documentation practices, and patterns of historical under-diagnosis. As a result, predictions may be systematically less accurate or less well-calibrated for underrepresented populations or for clinical settings that diverge from the training distribution. Automation bias is a further concern: clinicians may over-rely on model outputs, particularly for rare conditions where confidence is poorly calibrated, or for tasks where model predictions do not correspond to an established clinical use of EEG. We therefore view CLEF as a tool to augment, not replace, expert interpretation. Prospective validation, subgroup fairness auditing, careful communication of uncertainty, and clear delineation of intended use are prerequisites to any clinical deployment.

\section{LLM-based Preprocessing Details}
\label{sec:llm-use}
\subsection{LLM Summarizer}
\label{sec:llm}
\begin{table}[h]
\centering
\scriptsize
\caption{\textbf{Clinical question set used for QA-consistency scoring.} Boolean and integer answers are scored by exact match ($\{0,1\}$); free-text answers are scored by LLM-judged similarity ($[0,1]$).}
\vspace{2mm}
\label{tab:qa_questions}
\renewcommand{\arraystretch}{1.15}
\setlength{\tabcolsep}{6pt}
\begin{tabular}{p{0.78\linewidth} p{0.13\linewidth}}
\toprule
Question & Scoring \\
\midrule
Is there a mention of sleep in the EEG? & Exact \\
Is the EEG classified as abnormal? & Exact \\
Is there a mention of generalized slowing? & Exact \\
\quad If yes, what is the frequency? & Exact \\
Is there a mention of a posterior dominant rhythm? & Exact \\
\quad If yes, what is the frequency? & Exact \\
Is there a mention of diffuse beta activity? & Exact \\
Is there epileptiform activity present (seizure, spikes, discharges, etc.)? & Exact \\
Is there a mention of burst suppression? & Exact \\
Is there a mention of a normal anterior--posterior (AP) gradient? & Exact \\
Is there a mention of background activity being symmetric between hemispheres? & Exact \\
Is there a mention of vertex sharp waves? & Exact \\
Are there any sleep spindles present? & Exact \\
Is there a mention of K-complexes? & Exact \\
Is there a mention of positive occipital sharp transients of sleep (POSTS)? & Exact \\
Are spike and sharp waves present? & Exact \\
Is there a mention of generalized periodic discharges? & Exact \\
Is there a mention of lateralized periodic discharges? & Exact \\
Is there a mention of focal slowing? & Exact \\
\quad If yes, describe the focal slowing. & Similarity \\
What medications are mentioned in the report? & Similarity \\
What pre-existing conditions are mentioned in the report? & Similarity \\
What potential diagnoses were mentioned? & Similarity \\
List suggested medications in the report (mentioned but not currently taken by the patient). & Similarity \\
\bottomrule
\end{tabular}
\end{table}

We preprocess each report with an LLM summarizer before alignment. The summarization prompt and output token length materially affects how much clinical content is preserved, so we select both via the automated QA-consistency pipeline. The question set used in the pipeline is listed in Table~\ref{tab:qa_questions}. Qwen-32B~\cite{qwen2, qwen2.5} is used for the LLM.

\textbf{Prompt candidates.} Candidate summarization prompts were generated by an LLM under guidance from EEG specialists, who provided reference examples and specified that prompts should prioritize clinically salient content (abnormalities, background activity, epileptiform features, sleep architecture) over boilerplate (electrode placements, timestamps, procedural notes). Table~\ref{tab:prompt_ablation} reports QA-consistency scores for several example candidates against best- and worst-case bounds: the best-case bound pairs each raw report with itself (perfect content preservation), while the worst-case bound replaces each report with content sampled from other patients (capturing only base-rate co-occurrence of findings across the corpus). The top-scoring prompt (\#1) is used in subsequent training.

{%
\tiny
\begin{longtable}{p{0.02\linewidth} >{\ttfamily}p{0.86\linewidth} p{0.03\linewidth}}
\caption{\textbf{Candidate summarization prompts evaluated by QA-consistency scoring at a fixed token length of 512.} Reference bounds are included: \emph{best} compares the full-length raw report against itself, while \emph{worst} replaces the report with content sampled from other patients' reports.}
\label{tab:prompt_ablation} \\
\toprule
ID & \textnormal{Prompt} & Score \\
\midrule
\endfirsthead

\multicolumn{3}{c}{\tablename\ \thetable\ -- \textit{Continued from previous page}} \\
\toprule
ID & \textnormal{Prompt} & Score \\
\midrule
\endhead

\midrule
\multicolumn{3}{r}{\textit{Continued on next page}} \\
\endfoot

\bottomrule
\endlastfoot

best  & \textnormal{Raw report compared against itself (upper bound).} & 0.905 \\
\midrule
\#1 & Convert this EEG report into dense, semantically-rich plain text optimized for embedding. Structure the output as continuous text ordered by diagnostic importance: \newline
1.Lead with primary findings and abnormalities using precise medical terminology \newline
2.Follow with background activity, rhythms, and patterns \newline
3.Include clinical correlations and interpretations \newline
4.Add technical quality notes \newline
5.End with intracranial EEG findings if present \newline
Remove: timestamps, electrode placements, redundant statements, formatting characters, comparison statements \newline
Preserve: exact medical terminology, negations (e.g., "no epileptiform activity"), qualifiers (e.g., "mild", "moderate"), seizure events, medications affecting recording, study type/duration \newline
Output: continuous plain text with complete sentences, front-loaded with diagnostically significant terms
& \textbf{0.871} \\
\midrule
\#2 &
Reformat this EEG report as plain text for text encoder processing: \newline
IMPORTANCE ORDER: \newline
- Scalp EEG abnormalities and diagnostic findings \newline
- Background activity description  \newline
- Clinical interpretation \newline
- Study quality and technical notes \newline
- Intracranial EEG (de-prioritized) \newline
EXCLUDE: timestamps, electrode maps, formatting, redundancies \newline
INCLUDE: exact medical terminology, all negations ("no", "absence of"), all qualifiers ("mild", "frequent"), medications, seizure markers, study duration/type \newline
STYLE: dense descriptive sentences, key diagnostic terms front-loaded, continuous plain text
& 0.866 \\
\midrule
\#3 &
Transform this EEG report into plain text for embedding by executing these steps sequentially: \newline
STEP 1 - Extract in this exact order: \newline
   a) Abnormal findings: epileptiform activity, seizures, focal slowing, asymmetries \newline
   b) Background activity: dominant frequency, organization, reactivity \newline
   c) Clinical interpretation: significance, correlation with symptoms \newline
   d) Technical notes: study quality, artifacts, duration, study type \newline
   e) Intracranial findings: (only if present, place last) \newline
STEP 2 - Remove these elements: \newline
   - All timestamps and time references \newline
   - Electrode placement descriptions (e.g., "10-20 system", "F3-C3") \newline
   - Formatting characters (\*, \#, bullets, line breaks, tabs) \newline
   - Repeated information across sections \newline
   - Comparison statements to prior studies \newline
STEP 3 - Preserve mandatory elements: \newline
   - Exact medical terminology (unchanged) \newline
   - All negations: "no", "absence of", "without" \newline
   - All qualifiers: "mild", "moderate", "severe", "occasional", "frequent" \newline
   - Medications mentioned during recording \newline
   - Event markers and seizure timestamps (event content only, not absolute times) \newline
STEP 4 - Output format: Dense continuous prose, complete sentences, diagnostically critical terms in first 50 words
& 0.859 \\
\midrule
\#4 &
Rewrite this EEG report as optimized plain text using these directives:
 \newline
ORDERING DIRECTIVE - Arrange content in exactly this sequence: \newline
1. Begin with: seizures, epileptiform discharges, focal abnormalities, significant slowing \newline
2. Continue with: background rhythm description, frequency, amplitude, organization, symmetry \newline
3. Then add: sleep stages, reactivity, activation results \newline
4. Then add: clinical impression, diagnostic significance \newline
5. Then add: study type, duration, quality indicators \newline
6. End with: intracranial EEG data (scalp EEG takes precedence) \newline
DELETION DIRECTIVE - Eliminate: \newline
Timestamps | Electrode names/positions | Format markup | Duplicate phrases | Prior study references \newline
PRESERVATION DIRECTIVE - Keep intact: \newline
Medical terminology (exact) | Negations (no/without/absence) | Quantifiers (mild/moderate/severe/frequent/rare) | Medications | Clinical context \newline
COMPOSITION DIRECTIVE - Write as: \newline
Continuous plain text. Dense sentences. Front-load critical findings. No special formatting. Prioritize semantic content over grammatical perfection.
& 0.837 \\
\midrule
\#5 &
You are an EEG specialist preparing a baseline summary from a prolonged continuous EEG recording, including intracranial data when present. 
Generate a single report that includes only findings that are stable or generally applicable across the recording, and explicitly exclude: 
Seizure narratives or electroclinical descriptions Event-specific timelines or examples Medication schedules Monitoring methodology or 
procedural boilerplate The report should be written so that the most clinically critical information appears first, followed by progressively 
lower-impact contextual details, such that any partial output still conveys the key diagnostic conclusions. If intracranial EEG details exist, 
give them lower priority than Scalp EEG Clearly state whether the EEG is normal or abnormal at the outset. If abnormal, specify the localization, 
lateralization, and frequency of interictal epileptiform abnormalities using formal clinical EEG terminology. Describe relevant background features, 
including organization, symmetry, focal or generalized slowing, and differences between scalp and intracranial recordings when applicable. 
Include physiologic state-related findings (drowsiness and sleep architecture), even if normal. Retain regional specificity (lobar and sublobar where 
appropriate) and concise frequency qualifiers (e.g., frequent, occasional, rare). Use formal, information-dense clinical EEG language, grouping findings 
by relevance rather than by rigid templates or timestamps. Do not include dates, example times, or references to specific events.
& 0.834 \\
\midrule
\#6 &
Extract and reorganize this EEG report into plain text for neural embedding. Prioritize information by semantic and diagnostic value: \newline
HIGHEST PRIORITY: abnormal findings, epileptiform discharges, seizure activity, significant background abnormalities \newline
MEDIUM PRIORITY: normal background characteristics, sleep architecture, reactivity, activation procedure results   \newline
LOWER PRIORITY: technical quality, artifact notes, study parameters \newline
LOWEST PRIORITY: intracranial EEG data \newline
Strip out: timestamps, electrode positions, duplicate information, formatting symbols, prior study comparisons \newline
Retain: precise medical terms, negations and qualifiers, medications, clinical context, event markers \newline
Format: dense continuous prose, diagnostically important content first
& 0.796\\
\midrule
\#7 &
You are summarizing the stable baseline EEG features from a prolonged continuous EEG recording.
Begin with a clear overall impression (normal vs abnormal), followed by the most clinically consequential abnormalities.
Findings should be ordered by diagnostic impact, not by report structure.
Describe interictal epileptiform abnormalities (or their absence) with localization, lateralization, and frequency using formal EEG terminology.
Include relevant background characteristics, slowing, sleep features, and qualitative evolution across the recording.
Intracranial EEG findings should be mentioned briefly only if they provide additional localization or clarification beyond scalp EEG, and should be clearly subordinated.
Interpretation is acceptable but must remain grounded in EEG findings.
Do not include seizures, examples, times, medications, artifacts, or procedural details.
& 0.783\\
\midrule
worst & \textnormal{Content sampled from other patients' reports (lower bound).} & 0.568 \\
\bottomrule
\end{longtable}
}%

\textbf{Token length.} Holding the selected prompt fixed, we ablate the output token length (Table~\ref{tab:token_length}). Performance peaks at 256 tokens; shorter outputs lose clinical detail, while longer outputs reintroduce boilerplate without improving recoverable content. We adopt 256 tokens as the summarization length for our training.

\begin{table}[h]
\centering
\footnotesize
\caption{\textbf{QA-consistency score under varying summarization token length.}}
\vspace{2mm}
\label{tab:token_length}
\begin{tabular}{cc}
\toprule
Token Length & QA-consistency score \\
\midrule
128 & 0.841 \\
256 & \textbf{0.875} \\
512 & 0.871 \\
\bottomrule
\end{tabular}
\end{table}

\subsection{Benchmark Processing Details}
\label{sec:benchmark-processing}
\textbf{EHR Tables.} Medication labels and disease labels were extracted from Electronic Health Record data to create the proposed benchmark. The medication names from the EHR data were processed and converted to ATC codes using LLMs and RxNorm ATC code API. Medication extraction was split into in-patient and out-patient settings. For EEG recordings during in-patient setting, only medication events that were completed within 1 day of the recording were included as positive labels. For out-patient setting, any medication prescriptions that were active at the time of the recording were included as positive labels. Importantly, we exclude all medication prescriptions that were ``PRN'', or ``take as needed''.  
Disease label extraction was split into chronic and non-chronic diseases. Chronic diseases could appear at any point in the EHR prior to the EEG recording, while non-chronic diseases need to have appeared at most one week prior to the recording. ICD-10 codes were sourced from the encounter diagnoses and patient problem lists, and at least two encounter diagnoses with the same ICD-10 code (on different days) or one from the problem list needed to appear prior to the session recording to satisfy a positive label. The ICD-10 codes were then grouped into Phecodes to represent clinically significant phenotypes. 
Downstream tasks with fewer than 15 positive labels in the test or validation set were discarded. 
 
\textbf{Clinical Reports.} Clinical reports also contained information regarding EEG features, medications, diseases and diagnoses. To extract features from the clinical report, an LLM was prompted to answer whether or not certain features were contained in the report, as follows:
\textit{"According to the report, does the EEG session contain XXX ? If report says it is present, answer 1. If report says it is not present, answer 0. If report does not mention it or unsure, answer -1. Please provide the reasoning for your answer."}
Cases where an event is not mentioned are discarded, to preserve negatives class for when report explicitly states that a feature is not present. Hermes 2 Pro tuned version of Llama-3-8B~\cite{Hermes-2-Pro-Llama-3-8B} was used for the question answering, to enforce structured output. 
Medications, diseases, and diagnoses were extracted from the report in list format using a prompt as follows: 
\textit{"List all XXX mentioned in the report. Do not explain, just list. If none are mentioned, answer none."} The results were then filtered, processed, and grouped with an LLM into Phecodes similar to the EHR disease labels. In the case where tasks from the clinical reports were overlapping with those from the EHR, the clinical report based downstream was preferred. 

\section{Model and Training Details}
\label{sec:model-training-details}
\textbf{EEG Spectrogram.}
Signals sampled at 200\,Hz were bandpass-filtered between 0.1 and 75\,Hz with a zero-phase IIR filter, and line noise was suppressed with notch filters at 60\,Hz and harmonics up to the Nyquist frequency. 
Each channel was converted to a time--frequency representation via multitaper STFT using a 4-second window (800 samples) with a stride of 0.625\,s (125 samples). 
Spectral concentration was controlled by DPSS tapers with time-halfbandwidth product $NW = 2.0$, retaining $K = 3$ tapers under a low-bias criterion (eigenvalue $> 0.9$). 
The multitaper power spectrum was computed as the eigenvalue-weighted average of squared FFT magnitudes across tapers, normalized by taper weights and sampling rate, then converted to decibels via $10\log_{10}P$. Frequency bins in $[0, 32)$\,Hz at a resolution of 0.25\,Hz yield $H = 128$ bins; for a 1280\,s session the stride produces $W = 2048$ frames. Spectrogram values are linearly normalized from $[-40, 40]$\,dB to $[-1, 1]$, with out-of-range values hard-clamped.

\textbf{EEG Tokenizer.}
The tokenizer comprises a convolutional stem (38 input channels: 19 EEG spectrograms concatenated with their 19-channel mask) followed by five levels of residual blocks with progressive downsampling, yielding total spatial downsampling of $16\times$ in frequency and $8\times$ in time. 
The output is  quantized against a codebook of $K=4096$ entries. The de-tokenizer mirrors the tokenizer in reverse, reconstructing the full 19-channel spectrogram from the quantized latent. 
A PatchGAN discriminator (3 strided convolutional layers) provides adversarial supervision.
The adversarial weight $\lambda_{\text{adv}}$ is computed adaptively per batch as the ratio of the reconstruction to adversarial gradient norms. 
The differential reconstruction weight is $\gamma_{\text{diff}} = 4.0$ and the codebook loss weight is $\lambda_{\text{code}} = 0.8$ with commitment weight $\lambda_{\text{commit}} = 0.2$.
To let it expose to partial-montage conditions, with a probability $p_{\text{PSG}}=0.3$ we restrict the input to the PSG subset; independently, we mask each remaining channel with probability $p_{\text{drop}}=0.1$. 
The channel masking probabilities ramp linearly from 0 to their target values over the first 100{,}000 steps. The model is trained for 200{,}000 steps with the Adam optimizer ($\beta_1 = 0.0$, $\beta_2 = 0.99$) at a base learning $4.5 \times 10^{-6}$ (effective learning rate of $1.44 \times 10^{-4}$ for batch size 32), taking approximately 20 hours on 4 NVIDIA L40S GPUs.

\textbf{EEG Encoder.} The EEG encoder is a Vision Transformer (ViT). We experiment with three model scales: CLEF-Small (depth 4, hidden dimension 512, 8 attention heads), CLEF-Base (depth 12, hidden dimension 512, 8 attention heads), and CLEF-Large (depth 20, hidden dimension 768, 12 attention heads), all using MLP ratio 4, pre-norm Transformer blocks with GELU activation, flash attention, and dropout 0.1. 
Each 1280-second EEG recording is tokenized into a grid of discrete VQGAN tokens and simultaneously patchified, where each raw spectrogram patch is linearly projected and added to its corresponding token embedding, yielding sequences of 2048 tokens.
Shorter sessions are right-padded and masked from self-attention and reconstruction loss.
A learnable proxy token is prepended to the sequence and given its own positional embedding; its projected representation is later used in the decoder to stand in for dropped and masked positions.
During training, tokens are masked following a truncated Gaussian distribution over $[0.25, 1.0]$ with $\mu = 0.55$ and $\sigma = 0.15$; $r_{\text{drop}} = 0.25$ of tokens are dropped, while the remaining masked tokens are replaced with a learnable mask embedding. The encoder output is aggregated via mean pooling over all valid kept tokens.

\textbf{EEG Decoder.} The decoder is a Transformer with depth 4, hidden dimension 512, 8 attention heads, MLP ratio 4, and pre-norm blocks with GELU activation and dropout 0.1. Both dropped and masked tokens are replaced by the projected proxy token representation in the decoder, and all positions are given a learnable full-sequence positional embedding. 

\textbf{Masked Reconstruction.} A token prediction head (linear $\to$ GELU $\to$ LayerNorm $\to$ weight-tied projection against the token embedding matrix) maps decoder outputs to logits over the tokenizer codebook. The reconstruction loss is label-smoothing cross-entropy (smoothing $= 0.1$) computed only over the masked positions. The model is optimized with AdamW ($\beta_1 = 0.9$, $\beta_2 = 0.95$, weight decay $= 0.1$) at a base learning rate of $1 \times 10^{-3}$ (effective learning rate $5 \times 10^{-4}$ for batch size 128), trained for 10 epochs taking approximately 10 hours on 4 NVIDIA L40S GPUs. EMA-smoothed weights (decay $= 0.999$) are used for the next stage.

\textbf{Report Encoder.} The report encoder processes the summarized clinical EEG reports using a two-stage architecture. First, a frozen T5-v1.1-Base encoder ($768$-dimensional hidden states) extracts token-level features from reports tokenized to a maximum length of $256$ tokens, which receives no gradient updates during training.  The resulting sequence is linearly projected to the shared embedding dimension of $512$, augmented with learnable positional embeddings, and then refined by a $4$-layer Transformer encoder with $8$ attention heads, pre-norm layer normalization, and GELU activations. A final layer norm is applied, and the sequence is aggregated into a single $512$-dimensional embedding via mean pooling over non-padding tokens.

\textbf{EHR Encoder.} The EHR encoder represents patient clinical metadata as a variable-length token sequence fed into a Transformer.
Each patient's demographics are embedded independently via three learnable embedding tables: age is discretized into 10 decade-wide bins (0--9 through 90--100 years) plus one N/A token (11 entries total); sex has 3 categories (F, M, Unknown); and race/ethnicity has 8 categories (American Indian or Alaska Native, Asian, Black or African American, Multiracial, Native Hawaiian or Other Pacific Islander, Other Race, White, Unavailable).
Conditions and medications are each represented by a single learnable embedding table, covering 178 disease categories and 205 medication categories respectively.
For each patient, the condition and medication sequences consist of the embeddings whose corresponding categories are positive or active at the time of recording.
Condition and medication sequences are truncated and zero-padded to fixed lengths of 30 and 50 tokens respectively.
The resulting 83 tokens (3 demographic, 30 condition, 50 medication tokens) are concatenated and passed through a 4-layer Transformer encoder with 8 attention heads, pre-norm layer normalization, and GELU activations. A final layer norm is applied, and mean pooling over non-padded tokens produces a single 512-dimensional EHR embedding.

\textbf{Cross-Modal Alignment.} Each encoder's output is projected to a shared 512-dimensional embedding space via a linear projection, with separate projection matrices for the EEG--report and EEG--EHR alignment streams. Contrastive similarities are scaled by a fixed temperature $\tau = 0.07$. 
Recordings without a linked report are masked out from the report alignment loss $\mathcal{L}_{\text{Report}}$ for both the positives and negatives.
EEG tokens are randomly dropped at rate $r_{\text{drop}} = 0.25$ before being passed to the EEG encoder, reducing sequence length during training and acting as an implicit regularizer. The model is optimized with AdamW ($\beta_1 = 0.9$, $\beta_2 = 0.95$, weight decay $= 0.1$) at a base learning rate of $1 \times 10^{-3}$ (effective learning rate $5 \times 10^{-4}$ for batch size 128), with cosine decay following a 1-epoch linear warmup. Training runs for 20 epochs, taking approximately 20 hours on 4 NVIDIA L40S GPUs. EMA-smoothed weights (decay $= 0.9999$) are used at inference.

\textbf{Downstream Probing.} The encoder is frozen and its output is pooled across time to create a single 512-dimensional (or 768-dimensional for CLEF-Large) embedding. This embedding is then probed using a 2 layer MLP with intermediate dimension of size 1536 for 5 epochs, with learning rate of $1 \times 10^{-4}$. Each task has an independent probing head. During training, a random 1280-second segment was sampled from a file. During validation, the session was chunked into windows of size 1280 seconds, and all encoder outputs were pooled across sessions to yield a single embedding to probe. The model is optimized with AdamW, with weight decay $= 0.01$. Best validation epoch was chosen for each task and then evaluate on. All baselines were pooled and probed in identical manner. Each task was repeated 4 times and standard deviations were calculated and provided for each. 

\section{Ablation Experiments}
\label{sec:ablation}
We organize ablations around the design choices introduced in Section~\ref{sec:method}, evaluated on averaged session-level downstream AUROC on our validation set. Three evaluation settings are reported throughout: Recon-Only uses the Stage~I encoder directly; Align-Only trains Stage~II from a randomly-initialized encoder; Recon$\to$Align is the full sequential pipeline.

\subsection{Input}

\textbf{Spectrogram vs.\ raw time-series input (Table~\ref{tab:input-diff}).}
Substituting a time-series input for the spectrogram degrades every downstream evaluation. The gap is most pronounced in Recon-Only (0.5740 vs.\ 0.6261): raw time-domain signals over windows of several seconds are too high-entropy for the model to reconstruct meaningfully, and the failed reconstruction pretraining provides no benefit to the subsequent Align stage (Recon$\to$Align 0.7171 vs.  Align-Only: 0.7203).

\textbf{Token vs.\ pixel reconstruction targets (Table~\ref{tab:input-diff}).}
Token-based reconstruction yields 0.6759 on the Recon-Only setting, a substantial improvement over spectrogram-pixel reconstruction at 0.6261, confirming that the discrete-codebook bottleneck cleans the reconstruction target.
Also, token-based pretraining provides larger boost for Recon$\to$Align compared to Align-Only (0.7134 $\to$ 0.7384, a gain of 0.0250), far exceeding that of pixel-based pretraining (0.7311 $\to$ 0.7386, a gain of 0.0075). 

\textbf{Token + PatchEmbed input (Table~\ref{tab:input-diff}).}
Tokens excel at reconstruction but are the \emph{weakest} input for Align-only training, since quantization discards information that the contrastive objective could otherwise exploit. ``Token + PatchEmbed'' recovers the details and achieves the best score on the full Recon$\to$Align setting.
Notably, on Recon-Only, Token + PatchEmbed (0.6536) underperforms Token alone (0.6759). We attribute this to a shortcut effect: the VQGAN tokenizer computes each discrete code from a spatial context, so the continuous patches co-located with a masked position can leak information about its token target, making local reconstruction easier and reducing pressure on the encoder to learn globally predictive representations.

\begin{table}[h]
    \footnotesize
    \centering
        \caption{\textbf{Ablation of input-level design.} Best result per setting in \textbf{bold}.}
    \vspace{2mm}
    \begin{tabular}{ccccc}
    \toprule
         & Recon-Only & Align-Only & Recon$\to$Align\\
    \midrule
     Time-Series  & 0.5740 & 0.7203 & 0.7171 \\
     Spectrogram & 0.6261 & 0.7311 & 0.7386\\
     Token  & \textbf{0.6759} & 0.7134  & 0.7384\\
     Token + PatchEmbed & 0.6536 & \textbf{0.7355} & \textbf{0.7472}\\
    \bottomrule
    \end{tabular}
    \label{tab:input-diff}
\end{table}

\subsection{Tokenization}

\textbf{Joint vs.\ per-channel tokenization (Table~\ref{tab:vqgan}).}
Joint mutli-channel tokenization improves reconstruction correlation in every frequency band over per-channel tokenization, despite using $2.375\times$ fewer tokens\footnote{The per-channel tokenizer is trained with $8\times$ finer downsampling of the time--frequency grid to keep the overall token budget comparable ($2.375\times$).}, confirming that modeling inter-channel dependencies jointly is more effective than treating channels independently.

\textbf{Mask scheduler (Table~\ref{tab:vqgan}).}
Removing the linear mask scheduler (``w/o Mask Scheduler'') and applying the full masking rate from the start of training degrades final reconstruction quality, consistent with the hypothesis that early training is otherwise spent fighting an ill-posed imputation problem before useful spatial structure has been learned.

\textbf{Inter-channel-emphasized loss (Tables~\ref{tab:vqgan} and~\ref{tab:mean-diff}).}
The joint tokenizer trained with vanilla $\ell_1$ (``w/o Inter-Channel-Emphasized'') achieves high mean-correlation $r_{\text{mean}}{=}0.9695$ but a differential-correlation $r_{\text{diff}}{=}0.5697$ that is \emph{worse} than the per-channel's $0.6032$. This is because the per-channel tokenizer cannot trade differential fidelity for mean fidelity by construction, whereas the joint tokenizer under vanilla $\ell_1$ actively collapses inter-channel variation in pursuit of common-mode accuracy. With the decomposed loss, our full model reaches $r_{\text{diff}}{=}0.6036$ (above the per-channel baseline) at the cost of $r_{\text{mean}}$ dropping marginally to $0.9619$. Table~\ref{tab:mean-diff} confirms this trade translates into downstream gains across all settings.

\begin{table}[h]
    \footnotesize
    \centering
        \caption{\textbf{Tokenizer reconstruction quality under ablated configurations.} Per-sample Pearson correlation ($r$) averaged over the validation set. $r_{\text{mean}}$: cross-channel mean spectrogram; $r_{\text{diff}}$: per-channel deviation from the mean; $r_{\delta}$--$r_{\beta}$: individual frequency bands; $r_{\text{all}}$: all channels and bands jointly.}
    \vspace{3mm}
    \resizebox{\textwidth}{!}{%
    \begin{tabular}{cccccccccccc}
    \toprule
         & $r_{\text{all}}$ & $r_{\text{mean}}$ & $r_{\text{diff}}$ & $r_{\delta}$ & $r_{\theta}$ & $r_{\alpha}$  & $r_{\sigma}$  & $r_{\beta}$  \\
    \midrule
     Multi-Channel, w/ All                         & 0.8981 & 0.9619 & 0.6036 & 0.8042 & 0.7028 & 0.6893 & 0.6688 & 0.7518 \\
     Multi-Channel, w/o Mask Scheduler             & 0.8942 & 0.9604 & 0.5895 & 0.8005 & 0.6934 & 0.6780 & 0.6518 & 0.7391 \\
     Multi-Channel, w/o Inter-Channel-Emphasized   & 0.8984 & 0.9695 & 0.5697 & 0.8051 & 0.7024 & 0.6912 & 0.6713 & 0.7507 \\
     Per-Channel (2.4x tokens)                     & 0.8692 & 0.9232 & 0.6032 & 0.7372 & 0.6097 & 0.5896 & 0.5614 & 0.6880 \\
    \bottomrule
    \end{tabular}}
    \label{tab:vqgan}
\end{table}
\begin{table}[h]
    \footnotesize
    \centering
    \caption{\textbf{Downstream performance comparison between w/ and w/o Inter-Channel-Emphasized tokenizer.}}
    \vspace{1mm}
    \begin{tabular}{lcc}
    \toprule
         & w/o Inter-Channel-Emphasized & w/ Inter-Channel-Emphasized \\
    \midrule
    Recon-Only (Token)         & 0.6650 & 0.6759 \\
    Align-Only (Token)          & 0.7080 & 0.7134 \\
    Recon$\to$Align (Token)     & 0.7309 & 0.7384 \\
    Recon-Only (Token + PatchEmbed)   & 0.6501 &  0.6536 \\
    Recon$\to$Align (Token + PatchEmbed) &  0.7450   & 0.7472 \\
    \bottomrule
    \end{tabular}
    \label{tab:mean-diff}
\end{table}

For the following ablations, we use token-only input, since pre-computed tokens can be cached and reused across runs, substantially reducing the cost of each ablation.

\subsection{Masking Hyperparameters}
 We ablate the truncated-Gaussian masking ratio in Tables~\ref{tab:mage-mean} ($\mu$) and~\ref{tab:mage-std} ($\sigma$). Performance peaks at $\mu{=}0.55$ and degrades on either side, since at very high masking ratios the model has too little context to predict the masked tokens reliably, while at low ratios the prediction problem is too easy to drive useful representation learning. Sensitivity to $\sigma$ is mild, with $\sigma{=}0.15$ slightly preferred. 
\begin{table}[h]
    \footnotesize
    \centering
        \caption{\textbf{Ablation of masking ratio distribution mean ($\mu$).}}
    \vspace{1mm}
    \begin{tabular}{ccccc}
    \toprule
         & 0.35 & 0.55 & 0.75 & 0.95\\
    \midrule
     AUC  & 0.6697 & 0.6759 & 0.6693 & 0.6524 \\
    \bottomrule
    \end{tabular}
    \label{tab:mage-mean}
\end{table}
\begin{table}[h]
    \footnotesize
    \centering
        \caption{\textbf{Ablation of masking ratio distribution standard deviation ($\sigma$).}}
    \vspace{1mm}
    \begin{tabular}{ccccc}
    \toprule
         & 0.05 & 0.15 & 0.25 & 0.35\\
    \midrule
     AUC  & 0.6721 & 0.6759 & 0.6731 & 0.6697 \\
    \bottomrule
    \end{tabular}
    \label{tab:mage-std}
\end{table}

\subsection{Cross-Modal Alignment}

\textbf{Both alignment objectives contribute (Table~\ref{tab:clip}).}
Removing either $\mathcal{L}_{\text{Report}}$ or $\mathcal{L}_{\text{EHR}}$ drops AUC (0.7279 and 0.7264 respectively, vs.\ 0.7384 with both). The two losses actually help each other optimize: with both present, converged validation losses are lower than when either is trained alone ($\mathcal{L}_{\text{Report}}^{\text{Val}}$: 1.347 vs.\ 1.424; $\mathcal{L}_{\text{EHR}}^{\text{Val}}$: 3.110 vs.\ 3.154), suggesting that the two modalities provide complementary supervisory signal rather than competing for model capacity.
 
\textbf{Pretraining matters (Table~\ref{tab:clip}).}
Removing Stage~I (``w/o Stage~I'') drops AUC from 0.7384 to 0.7134, and the converged alignment losses are substantially worse (1.780 and 3.432 vs.\ 1.347 and 3.110). Masked modeling thus gives Stage~II a meaningfully better starting point than random initialization.
\begin{table}[h]
    \footnotesize
    \centering
    \caption{\textbf{Ablation of cross-modal alignment objectives.}  $\mathcal{L}_{\text{Report}}^{\text{Val}}$ and $\mathcal{L}_{\text{EHR}}^{\text{Val}}$ are the converged validation values of the two contrastive losses. ``All'' uses Stage~I pretraining and both alignment objectives.}
    \label{tab:clip}
    \vspace{3mm}
    \begin{tabular}{lccc}
    \toprule
         & AUC  & $\mathcal{L}_{\text{Report}}^{\text{Val}}$ & $\mathcal{L}_{\text{EHR}}^{\text{Val}}$ \\
    \midrule
     All                                & 0.7384 & 1.347 & 3.110 \\
     w/o Stage~I                        & 0.7134 & 1.780 & 3.432 \\
     w/o $\mathcal{L}_{\text{Report}}$  & 0.7279 & ---   & 3.154 \\
     w/o $\mathcal{L}_{\text{EHR}}$     & 0.7264 & 1.424 & ---     \\
    \bottomrule
    \end{tabular}
\end{table}
 
\subsection{Training Scheme}
 
\textbf{Sequential beats joint (Table~\ref{tab:train-scheme}).}
Training $\mathcal{L}_{\text{Recon}}$ and $\mathcal{L}_{\text{Align}}$ jointly from scratch yields 0.7226, noticeably below the 0.7384 of the sequential pipeline. Extending the joint run to longer training does not close the gap (0.7214), ruling out undertraining as the cause and supporting our interpretation that the two objectives interfere when optimized simultaneously from an uninitialized encoder.
 
\textbf{Retaining reconstruction in Stage~II does not help (Table~\ref{tab:train-scheme}).}
Keeping $\mathcal{L}_{\text{Recon}}$ alongside the alignment losses during Stage~II is tied with the sequential baseline (0.7381), and raising the masking ratio does not change the picture (0.7371, 0.7375). We therefore default to the sequential pipeline with no reconstruction term in Stage~II.

\begin{table}[h]
    \footnotesize
    \centering
    \caption{\textbf{Ablation of training scheme.} ``Recon$\to$Align (two-stage)'' is our default; ``Recon+Align (Stage~II)'' variants retain $\mathcal{L}_{\text{Recon}}$ during Stage~II at masking ratio $r$; ``Recon+Align (single-stage)'' jointly optimizes both objectives from scratch.}
    \label{tab:train-scheme}
    \vspace{3mm}
    \begin{tabular}{lc}
    \toprule
     Training Scheme    & AUC   \\
    \midrule
     Recon$\to$Align (two-stage, default)          & 0.7384 \\
     \midrule
     Recon$+$Align (Stage~II), $r=0.25$                     & 0.7381 \\
     Recon$+$Align (Stage~II), $r\sim \mathcal{U}(0.25, 0.50)$ & 0.7371 \\
     Recon$+$Align (Stage~II), $r\sim \mathcal{U}(0.25, 0.75)$ & 0.7375 \\
     \midrule
     Recon$+$Align (single-stage)                    & 0.7226 \\
     Recon$+$Align (single-stage), + $10$ epochs         & 0.7214 \\
    \bottomrule
    \end{tabular}
\end{table}

\clearpage

\begin{table*}[t]
\section{Top 5 Tasks Per Category}
\label{sec:top-5}
\centering
\fontsize{6}{7.2}\selectfont   
{\setlength{\tabcolsep}{0pt}%
\textbf{Top-5 Diseases per Category}\\[3pt]
\begin{tabular}{>{\raggedright\arraybackslash}p{70.16pt}@{\hspace{2pt}}>{\raggedleft\arraybackslash}p{55.0pt}@{\hspace{8pt}}>{\raggedright\arraybackslash}p{70.16pt}@{\hspace{2pt}}>{\raggedleft\arraybackslash}p{55.0pt}@{\hspace{8pt}}>{\raggedright\arraybackslash}p{70.16pt}@{\hspace{2pt}}>{\raggedleft\arraybackslash}p{55.0pt}}
    \toprule
    \multicolumn{2}{c}{\textbf{Crit. Illness}} & \multicolumn{2}{c}{\textbf{Syst. Infection}} & \multicolumn{2}{c}{\textbf{Cerebrovasc.}} \\
    \cmidrule(lr){1-2} \cmidrule(lr){3-4} \cmidrule(lr){5-6}
    \textit{Task} & \textit{AUROC} & \textit{Task} & \textit{AUROC} & \textit{Task} & \textit{AUROC} \\
    \midrule
    Respiratory failure & $0.907\pm0.003$ & Septic shock & $0.843\pm0.003$ & Cereb. vasospasm & $0.903\pm0.003$ \\
    Acute resp. failure & $0.905\pm0.003$ & Sars-CoV-2 & $0.827\pm0.003$ & Cerebral edema & $0.844\pm0.003$ \\
    Cardiac arrest & $0.895\pm0.002$ & Coronavirus & $0.827\pm0.004$ & Compression of brain & $0.817\pm0.007$ \\
    Cardiovasc. events & $0.879\pm0.005$ & Sepsis & $0.819\pm0.004$ & Intracranial hemor. & $0.781\pm0.004$ \\
    Post cardiac arrest & $0.832\pm0.003$ & Bact. Sepsis SIRS & $0.804\pm0.004$ & Idiopath. NPH & $0.768\pm0.003$ \\
    \bottomrule
  \end{tabular}
\par\vspace{4pt}
  \begin{tabular}{>{\raggedright\arraybackslash}p{70.16pt}@{\hspace{2pt}}>{\raggedleft\arraybackslash}p{55.0pt}@{\hspace{8pt}}>{\raggedright\arraybackslash}p{70.16pt}@{\hspace{2pt}}>{\raggedleft\arraybackslash}p{55.0pt}@{\hspace{8pt}}>{\raggedright\arraybackslash}p{70.16pt}@{\hspace{2pt}}>{\raggedleft\arraybackslash}p{55.0pt}}
    \toprule
    \multicolumn{2}{c}{\textbf{Neurodegeneration}} & \multicolumn{2}{c}{\textbf{Hepatic/Renal}} & \multicolumn{2}{c}{\textbf{Metabolic}} \\
    \cmidrule(lr){1-2} \cmidrule(lr){3-4} \cmidrule(lr){5-6}
    \textit{Task} & \textit{AUROC} & \textit{Task} & \textit{AUROC} & \textit{Task} & \textit{AUROC} \\
    \midrule
    Anoxic brain damage & $0.900\pm0.004$ & Alcoholic liver dis. & $0.824\pm0.003$ & Encephalopathy & $0.799\pm0.001$ \\
    Brain damage/death & $0.816\pm0.004$ & Acute kidney failure & $0.789\pm0.005$ & Acidosis & $0.788\pm0.003$ \\
    Alzheimer's disease & $0.768\pm0.006$ & Renal failure & $0.785\pm0.004$ & Mixed acid-base & $0.780\pm0.004$ \\
    Dementias & $0.737\pm0.002$ & Fibrosis/cirrhosis & $0.774\pm0.002$ & Toxic encephalopathy & $0.766\pm0.003$ \\
    Anoxic brain injury & $0.727\pm0.011$ & Cirrhosis of liver & $0.768\pm0.008$ & Hyperkalemia & $0.725\pm0.001$ \\
    \bottomrule
  \end{tabular}
\par\vspace{4pt}
  \begin{tabular}{>{\raggedright\arraybackslash}p{70.16pt}@{\hspace{2pt}}>{\raggedleft\arraybackslash}p{55.0pt}@{\hspace{8pt}}>{\raggedright\arraybackslash}p{70.16pt}@{\hspace{2pt}}>{\raggedleft\arraybackslash}p{55.0pt}@{\hspace{8pt}}>{\raggedright\arraybackslash}p{70.16pt}@{\hspace{2pt}}>{\raggedleft\arraybackslash}p{55.0pt}}
    \toprule
    \multicolumn{2}{c}{\textbf{Oncology}} & \multicolumn{2}{c}{\textbf{Neurodevelopmental}} & \multicolumn{2}{c}{\textbf{Subst. Use}} \\
    \cmidrule(lr){1-2} \cmidrule(lr){3-4} \cmidrule(lr){5-6}
    \textit{Task} & \textit{AUROC} & \textit{Task} & \textit{AUROC} & \textit{Task} & \textit{AUROC} \\
    \midrule
    Brain malignancy & $0.816\pm0.003$ & Cerebral palsy & $0.881\pm0.007$ & Opioid use disorders & $0.780\pm0.004$ \\
    Acute leukemia & $0.746\pm0.005$ & Autism & $0.744\pm0.008$ & Opioid abuse/depend. & $0.774\pm0.003$ \\
    Secondary malignancy & $0.733\pm0.003$ & Global dev. delay & $0.707\pm0.006$ & Psychoact. depend. & $0.712\pm0.002$ \\
    Benign brain tumor & $0.732\pm0.011$ & Tuberous sclerosis & $0.697\pm0.045$ & Stimulant use dis. & $0.712\pm0.004$ \\
    Non-Hodgkin lymphoma & $0.731\pm0.007$ & Brain malformations & $0.678\pm0.006$ & Stimulant abuse/dep. & $0.703\pm0.011$ \\
    \bottomrule
  \end{tabular}
\par\vspace{4pt}
  \begin{tabular}{>{\raggedright\arraybackslash}p{70.16pt}@{\hspace{2pt}}>{\raggedleft\arraybackslash}p{55.0pt}@{\hspace{8pt}}>{\raggedright\arraybackslash}p{70.16pt}@{\hspace{2pt}}>{\raggedleft\arraybackslash}p{55.0pt}@{\hspace{8pt}}>{\raggedright\arraybackslash}p{70.16pt}@{\hspace{2pt}}>{\raggedleft\arraybackslash}p{55.0pt}}
    \toprule
    \multicolumn{2}{c}{\textbf{Cogn./Lang.}} & \multicolumn{2}{c}{\textbf{Epilepsy}} & \multicolumn{2}{c}{\textbf{Psychiatric}} \\
    \cmidrule(lr){1-2} \cmidrule(lr){3-4} \cmidrule(lr){5-6}
    \textit{Task} & \textit{AUROC} & \textit{Task} & \textit{AUROC} & \textit{Task} & \textit{AUROC} \\
    \midrule
    Transient amnesia & $0.763\pm0.002$ & Partial epilepsy & $0.746\pm0.005$ & Delusional disorders & $0.736\pm0.008$ \\
    Coma & $0.754\pm0.003$ & Epilepsy & $0.742\pm0.001$ & Bipolar I disorder & $0.732\pm0.002$ \\
    Delirium syndrome & $0.752\pm0.006$ & Generalized epilepsy & $0.725\pm0.004$ & Primary psychosis & $0.730\pm0.006$ \\
    Altered mental status & $0.695\pm0.003$ & Recurrent seizures & $0.693\pm0.001$ & PTSD & $0.700\pm0.003$ \\
    Alt. consciousness & $0.682\pm0.006$ & -- & -- & Bipolar disorder & $0.695\pm0.005$ \\
    \bottomrule
  \end{tabular}
\par\vspace{4pt}
  \begin{tabular}{>{\raggedright\arraybackslash}p{70.16pt}@{\hspace{2pt}}>{\raggedleft\arraybackslash}p{55.0pt}@{\hspace{8pt}}>{\raggedright\arraybackslash}p{70.16pt}@{\hspace{2pt}}>{\raggedleft\arraybackslash}p{55.0pt}}
    \toprule
    \multicolumn{2}{c}{\textbf{Hematol./Imm.}} & \multicolumn{2}{c}{\textbf{Movement}} \\
    \cmidrule(lr){1-2} \cmidrule(lr){3-4}
    \textit{Task} & \textit{AUROC} & \textit{Task} & \textit{AUROC} \\
    \midrule
    Pancytopenia & $0.706\pm0.005$ & Parkinsonism & $0.713\pm0.003$ \\
    Thrombocytopenia & $0.706\pm0.006$ & Myoclonus & $0.687\pm0.006$ \\
    Aplastic anemia & $0.698\pm0.004$ & Tremor & $0.637\pm0.008$ \\
    Anemia & $0.672\pm0.003$ & Vertigo & $0.630\pm0.011$ \\
    ITP & $0.661\pm0.007$ & Paroxysmal vertigo & $0.622\pm0.024$ \\
    \bottomrule
  \end{tabular}
\par\vspace{4pt}

\medskip
\textbf{Top-5 Medications per Category}\\[3pt]
  \begin{tabular}{>{\raggedright\arraybackslash}p{70.16pt}@{\hspace{2pt}}>{\raggedleft\arraybackslash}p{55.0pt}@{\hspace{8pt}}>{\raggedright\arraybackslash}p{70.16pt}@{\hspace{2pt}}>{\raggedleft\arraybackslash}p{55.0pt}@{\hspace{8pt}}>{\raggedright\arraybackslash}p{70.16pt}@{\hspace{2pt}}>{\raggedleft\arraybackslash}p{55.0pt}}
    \toprule
    \multicolumn{2}{c}{\textbf{Analgesics}} & \multicolumn{2}{c}{\textbf{Antiepileptics}} & \multicolumn{2}{c}{\textbf{Anesthetics}} \\
    \cmidrule(lr){1-2} \cmidrule(lr){3-4} \cmidrule(lr){5-6}
    \textit{Task} & \textit{AUROC} & \textit{Task} & \textit{AUROC} & \textit{Task} & \textit{AUROC} \\
    \midrule
    Hydromorphone & $0.913\pm0.005$ & Clobazam & $0.877\pm0.001$ & Propofol & $0.925\pm0.001$ \\
    Fentanyl & $0.888\pm0.002$ & Fosphenytoin & $0.847\pm0.007$ & Dexmedetomidine & $0.898\pm0.002$ \\
    Acetaminophen & $0.877\pm0.002$ & Cannabidiol & $0.834\pm0.005$ & Ketamine & $0.806\pm0.008$ \\
    Morphine & $0.859\pm0.005$ & Zonisamide & $0.826\pm0.002$ & Bupivacaine & $0.700\pm0.009$ \\
    Methadone & $0.790\pm0.002$ & Pregabalin & $0.825\pm0.001$ & -- & -- \\
    \bottomrule
  \end{tabular}
\par\vspace{4pt}
  \begin{tabular}{>{\raggedright\arraybackslash}p{70.16pt}@{\hspace{2pt}}>{\raggedleft\arraybackslash}p{55.0pt}@{\hspace{8pt}}>{\raggedright\arraybackslash}p{70.16pt}@{\hspace{2pt}}>{\raggedleft\arraybackslash}p{55.0pt}@{\hspace{8pt}}>{\raggedright\arraybackslash}p{70.16pt}@{\hspace{2pt}}>{\raggedleft\arraybackslash}p{55.0pt}}
    \toprule
    \multicolumn{2}{c}{\textbf{Psycholeptics}} & \multicolumn{2}{c}{\textbf{Cardiovascular}} & \multicolumn{2}{c}{\textbf{Anti-infectives}} \\
    \cmidrule(lr){1-2} \cmidrule(lr){3-4} \cmidrule(lr){5-6}
    \textit{Task} & \textit{AUROC} & \textit{Task} & \textit{AUROC} & \textit{Task} & \textit{AUROC} \\
    \midrule
    Midazolam & $0.876\pm0.005$ & Amiodarone & $0.780\pm0.003$ & Ceftazidime & $0.790\pm0.002$ \\
    Suvorexant & $0.852\pm0.002$ & Nebivolol & $0.771\pm0.037$ & Cefepime & $0.783\pm0.005$ \\
    Ramelteon & $0.811\pm0.008$ & Epinephrine & $0.764\pm0.004$ & Meropenem & $0.762\pm0.003$ \\
    Baclofen & $0.808\pm0.005$ & Dopamine & $0.762\pm0.017$ & Pip.+tazobactam & $0.742\pm0.002$ \\
    Olanzapine & $0.790\pm0.008$ & Furosemide & $0.749\pm0.006$ & Aciclovir (high) & $0.736\pm0.011$ \\
    \bottomrule
  \end{tabular}
\par\vspace{4pt}
  \begin{tabular}{>{\raggedright\arraybackslash}p{70.16pt}@{\hspace{2pt}}>{\raggedleft\arraybackslash}p{55.0pt}@{\hspace{8pt}}>{\raggedright\arraybackslash}p{70.16pt}@{\hspace{2pt}}>{\raggedleft\arraybackslash}p{55.0pt}@{\hspace{8pt}}>{\raggedright\arraybackslash}p{70.16pt}@{\hspace{2pt}}>{\raggedleft\arraybackslash}p{55.0pt}}
    \toprule
    \multicolumn{2}{c}{\textbf{Metabolic}} & \multicolumn{2}{c}{\textbf{Antidepressants}} & \multicolumn{2}{c}{\textbf{Respiratory}} \\
    \cmidrule(lr){1-2} \cmidrule(lr){3-4} \cmidrule(lr){5-6}
    \textit{Task} & \textit{AUROC} & \textit{Task} & \textit{AUROC} & \textit{Task} & \textit{AUROC} \\
    \midrule
    Enoxaparin & $0.845\pm0.005$ & Trazodone & $0.805\pm0.007$ & Loratadine & $0.680\pm0.002$ \\
    Insulin (sol.) & $0.768\pm0.003$ & Sertraline & $0.728\pm0.010$ & Diphenhydramine & $0.647\pm0.011$ \\
    Metformin & $0.721\pm0.018$ & Citalopram & $0.707\pm0.005$ & Cetirizine & $0.589\pm0.002$ \\
    Insulin (isoph.) & $0.710\pm0.004$ & Mirtazapine & $0.691\pm0.003$ & -- & -- \\
    Estradiol & $0.701\pm0.014$ & Bupropion & $0.686\pm0.006$ & -- & -- \\
    \bottomrule
  \end{tabular}
}
\label{tab:categories}
\end{table*}

\clearpage
\section{Detailed Results}
\label{sec:detailed-results}
{%
\setlength{\tabcolsep}{0pt}%
\setlength{\arrayrulewidth}{0.3pt}%
\fontsize{4}{4.8}\selectfont

\begin{longtable}{@{}>{\fontsize{4}{4.8}\selectfont \raggedright\arraybackslash}p{71.0pt}@{\hspace{4pt}}>{\fontsize{4}{4.8}\selectfont \raggedright\arraybackslash}p{36.0pt}@{\hspace{4pt}}>{\fontsize{4}{4.8}\selectfont \raggedright\arraybackslash}p{18.0pt}@{\hspace{4pt}}>{\fontsize{4}{4.8}\selectfont \raggedleft\arraybackslash}p{12.0pt}@{\hspace{4pt}}>{\fontsize{4}{4.8}\selectfont \raggedleft\arraybackslash}p{12.0pt}@{\hspace{2pt}|\hspace{2pt}}>{\fontsize{4}{4.8}\selectfont \raggedright\arraybackslash}p{28.0pt}@{\hspace{2pt}|\hspace{2pt}}>{\fontsize{4}{4.8}\selectfont \raggedright\arraybackslash}p{28.0pt}@{\hspace{2pt}|\hspace{2pt}}>{\fontsize{4}{4.8}\selectfont \raggedright\arraybackslash}p{28.0pt}@{\hspace{2pt}|\hspace{2pt}}>{\fontsize{4}{4.8}\selectfont \raggedright\arraybackslash}p{28.0pt}@{\hspace{2pt}|\hspace{2pt}}>{\fontsize{4}{4.8}\selectfont \raggedright\arraybackslash}p{28.0pt}@{\hspace{2pt}|\hspace{2pt}}>{\fontsize{4}{4.8}\selectfont \raggedright\arraybackslash}p{28.0pt}@{\hspace{2pt}|\hspace{2pt}}>{\fontsize{4}{4.8}\selectfont \raggedright\arraybackslash}p{28.0pt}@{}}
\toprule
\textbf{Task} & \textbf{Subcategory} & \textbf{Category} & \textbf{N+} & \textbf{N-} & \textbf{CLEF~(Full)} & \textbf{CLEF~(Recon. Only)} & \textbf{BioT} & \textbf{REVE} & \textbf{CBraMod} & \textbf{LaBraM} & \textbf{NeuroLM} \\
\midrule
\endfirsthead

\midrule
\endhead

\midrule
\multicolumn{12}{r}{\textit{Continued on next page}} \\
\endfoot

\bottomrule
\endlastfoot

Overall downstream &  &  & --- & --- & \textbf{0.7373 ± 0.0062} & 0.6709 ± 0.0080 & 0.5585 ± 0.0282 & 0.6279 ± 0.0172 & 0.5649 ± 0.0171 & 0.6300 ± 0.0137 & 0.6459 ± 0.0120 \\
Propofol & Anesthetics & Medications & 1,865 & 12,687 & \textbf{0.9253 ± 0.0011} & 0.9049 ± 0.0037 & 0.7435 ± 0.0049 & 0.8791 ± 0.0016 & 0.7591 ± 0.0039 & 0.8505 ± 0.0048 & 0.8715 ± 0.0048 \\
Hydromorphone & Analgesics & Medications & 426 & 4,168 & \textbf{0.9132 ± 0.0051} & 0.7497 ± 0.0131 & 0.5207 ± 0.0384 & 0.7086 ± 0.0147 & 0.5855 ± 0.0337 & 0.7357 ± 0.0280 & 0.6813 ± 0.0229 \\
Respiratory failure & Crit.Ill./Cardiopulm. & Diseases & 1,763 & 11,242 & \textbf{0.9073 ± 0.0028} & 0.8865 ± 0.0027 & 0.6885 ± 0.0438 & 0.8512 ± 0.0060 & 0.7570 ± 0.0038 & 0.8311 ± 0.0015 & 0.8507 ± 0.0044 \\
Acute respiratory failure & Crit.Ill./Cardiopulm. & Diseases & 1,742 & 11,554 & \textbf{0.9054 ± 0.0032} & 0.8742 ± 0.0025 & 0.6767 ± 0.0315 & 0.8339 ± 0.0093 & 0.7527 ± 0.0078 & 0.8121 ± 0.0029 & 0.8308 ± 0.0056 \\
Cerebrovascular vasospasm and vasoconstriction* & Cerebrovasc./Struct. & Diseases & 770 & 6,565 & \textbf{0.9034 ± 0.0025} & 0.8692 ± 0.0153 & 0.6464 ± 0.0396 & 0.8392 ± 0.0074 & 0.6930 ± 0.0193 & 0.8272 ± 0.0064 & 0.8514 ± 0.0045 \\
Anoxic brain damage & Neurodegeneration & Diseases & 848 & 7,004 & \textbf{0.9005 ± 0.0038} & 0.8985 ± 0.0108 & 0.7257 ± 0.0196 & 0.8492 ± 0.0048 & 0.7515 ± 0.0073 & 0.7752 ± 0.0113 & 0.8280 ± 0.0063 \\
Dexmedetomidine & Anesthetics & Medications & 1,509 & 12,378 & \textbf{0.8978 ± 0.0023} & 0.8436 ± 0.0018 & 0.7530 ± 0.0210 & 0.8489 ± 0.0032 & 0.7499 ± 0.0018 & 0.8273 ± 0.0050 & 0.8422 ± 0.0039 \\
\bottomrule
\end{longtable}
}

\clearpage
\section{Robustness Analyses}
\label{sec:analysis}

We analyze how channel count and recording length affect representation quality. All ablations report mean AUROC averaged over the 234 benchmark tasks with the same seed.

\textbf{Channel Count.} Because downstream deployments may have access to fewer channels than the full 10--20 montage, we evaluate the pre-trained 42M model with progressively smaller channel subsets at probing time, masking out the unavailable channels: 1 ch (C4); 3 ch (F4, C4, O2), the AASM-recommended PSG montage~\cite{berry2017aasm}; 6 ch (F3, F4, C3, C4, O1, O2); 12 ch (Fp1, Fp2, F3, F4, C3, C4, P3, P4, O1, O2, T3, T4); and 19 ch (full 10--20 montage). As shown in Table~\ref{tab:channels}, performance rises sharply from 1 to 3 channels and from 3 to 6 channels, then plateaus, suggesting that most clinically relevant content is recoverable from the AASM PSG montage and that CLEF representations remain useful in low-channel deployments.

\begin{table}[h]
\centering
\footnotesize
\caption{\textbf{Impact of Channel Count.}}
\vspace{2mm}
\label{tab:channels}
\footnotesize
\begin{tabular}{lc}
\toprule
Channels Count & AUROC \\
\midrule
1   & 0.6392 \\
3   & 0.7106 \\
6   & 0.7311 \\
12  & 0.7325 \\
19  & 0.7354 \\
\bottomrule
\end{tabular}
\end{table}

\textbf{Recording Length.} To assess the effect of recording duration on patient representation quality, we trim recordings to various maximum lengths at probing time and re-evaluate the pre-trained 42M model. As shown in Table~\ref{tab:recording-length}, AUROC improves monotonically with available recording length, rising from 0.7065 at 6 minutes to 0.7354 with the full session, with gains largest at short durations. This pattern is consistent with the clinical practice of acquiring routine EEGs of approximately 20 minutes. It also indicates that CLEF representations degrade gracefully under short recordings, supporting the settings where full-length sessions are not available.

\begin{table}[h]
\centering
\footnotesize
\caption{\textbf{Impact of Recording Length.}}
\vspace{2mm}
\label{tab:recording-length}
\footnotesize
\begin{tabular}{lc}
\toprule
Recording Length & AUROC \\
\midrule
6 min   & 0.7065 \\
15 min   & 0.7203 \\
45 min   & 0.7285 \\
90 min  & 0.7310 \\
All  & 0.7354 \\
\bottomrule
\end{tabular}
\end{table}

\section{Assets and Licenses}
\label{sec:assets}
In this work, we utilized the following existing assets:

\textbf{Harvard Electroencephalography Database (HEEDB)~\cite{sun2025harvard}.} We used version 4.1, accessed via \url{https://bdsp.io/content/harvard-eeg-db/4.1/}. The data is provided under the BDSP Credentialed Health Data License, which restricts usage to non-commercial scientific research. 

\textbf{TUH EEG Corpus~\cite{obeid2016temple}.} We accessed the Temple University Hospital (TUH) EEG Corpus via \url{https://isip.piconepress.com/projects/tuh_eeg/}. The data is released by the Neural Engineering Data Consortium as an unencumbered open-source resource, free for both research and commercial use. 

\textbf{George B. Moody PhysioNet Challenge 2026~\cite{hsp}.} We accessed the 2026 Challenge resources via \url{https://moody-challenge.physionet.org/2026/}. The Challenge data are sourced from the Human Sleep Project (HSP) database (v2.0), a collection of clinical polysomnography (PSG) recordings hosted on the Brain Data Science Platform (\url{https://bdsp.io/content/hsp/2.0/}). As stated by the Challenge organizers, the license and access conditions for the data are inherited from the Human Sleep Project, which is distributed under the BDSP Credentialed Health Data License and restricts usage to non-commercial scientific research.

\textbf{MNE-Python~\cite{gramfort2013meg}.} MNE-Python is distributed under the BSD-3-Clause License. We accessed it via \url{https://github.com/mne-tools/mne-python}.

\textbf{Braindecode~\cite{braindecode,HBM:HBM23730}.} The Braindecode toolbox is primarily distributed under the BSD-3-Clause License (some auxiliary components are licensed under CC BY-NC 4.0). We accessed it via \url{https://github.com/braindecode/braindecode}.

\textbf{Qwen2.5-32B-Instruct~\cite{qwen2,qwen2.5}.} The model weights are distributed under the Apache License 2.0. We accessed the instruction-tuned 32B variant of the Qwen2.5 series via the Hugging Face Transformers library (\url{https://huggingface.co/Qwen/Qwen2.5-32B-Instruct}).

\textbf{T5-base Encoder~\cite{raffel2020exploring}.} The code and model weights are distributed under the Apache License 2.0. We accessed version 1.1 of the weights via the Hugging Face Transformers library (\url{https://huggingface.co/google-t5/t5-base}).

\textbf{Llama-3.1-70B~\cite{grattafiori2024llama}.} The model weights are distributed under the Llama 3.1 Community License (\texttt{llama3.1}). We accessed the weights via the Hugging Face Transformers library (\url{https://huggingface.co/meta-llama/Llama-3.1-70B}).

\textbf{Hermes-2-Pro-Llama-3-8B~\cite{Hermes-2-Pro-Llama-3-8B}.} The model weights are distributed under the Meta Llama 3 Community License (\texttt{llama3}), as the model is a fine-tune of Meta-Llama-3-8B. We accessed the weights via the Hugging Face Transformers library (\url{https://huggingface.co/NousResearch/Hermes-2-Pro-Llama-3-8B}).

\end{document}